\documentclass[sigconf, review=false, balance=true]{acmart}
\acmSubmissionID{217}

\setcitestyle{nosort,square} % nosort to allow for manual chronological ordering

\usepackage{pifont}
\usepackage{appendix}
\usepackage{siunitx}
\usepackage[capitalize]{cleveref}
\usepackage[utf8]{inputenc} % allow utf-8 input
\usepackage[T1]{fontenc}    % use 8-bit T1 fonts
\usepackage{url}            % simple URL typesetting
\usepackage{amsfonts}       % blackboard math symbols
\usepackage{nicefrac}       % compact symbols for 1/2, etc.
\usepackage{microtype}      % microtypography
\usepackage{wrapfig}
\usepackage{multirow}
\usepackage{listings}
\usepackage{color}
\usepackage{mathtools}
\usepackage{colortbl}
\usepackage{makecell}
\usepackage{soul}
\usepackage[ruled]{algorithm2e} % For algorithms

\copyrightyear{2024}
\acmYear{2024}
\setcopyright{rightsretained}
\acmConference[SIGGRAPH Conference Papers '24]{Special Interest Group on Computer Graphics and Interactive Techniques Conference Conference Papers '24}{July 27-August 1, 2024}{Denver, CO, USA}
\acmBooktitle{Special Interest Group on Computer Graphics and Interactive Techniques Conference Conference Papers '24 (SIGGRAPH Conference Papers '24), July 27-August 1, 2024, Denver, CO, USA}
\acmDOI{10.1145/3641519.3657402}
\acmISBN{979-8-4007-0525-0/24/07}

\definecolor[named]{yellow}{rgb}{1,1, 0.6}
\definecolor[named]{lightyellow}{rgb}{1,1, 0.8}
\definecolor[named]{orange}{rgb}{1, 0.8, 0.6}
\definecolor[named]{red}{rgb}{1, 0.6, 0.6}
\definecolor[named]{wincolor}{rgb}{0.85, 0.0, 0.0}
\definecolor[named]{darkyellow}{rgb}{0.8, 0.8, 0.5}
\definecolor[named]{darkred}{rgb}{0.7, 0.3, 0.3}
\definecolor[named]{darkgreen}{rgb}{0.3, 0.7, 0.3}
\definecolor[named]{blue}{rgb}{0, 0, 1.0}
\definecolor[named]{green}{rgb}{0, 1.0, 0}
\definecolor[named]{pink}{rgb}{1, 0.4, 0.7}

\let\originalleft\left
\let\originalright\right
\renewcommand{\left}{\mathopen{}\mathclose\bgroup\originalleft}
\renewcommand{\right}{\aftergroup\egroup\originalright}
\newcommand{\norm}[1]{\left\lVert#1\right\rVert}

\newcommand{\sphere}{\mathbf{\mathcal{S}}}

\newcommand{\rayorigin}{\mathbf{o}}
\newcommand{\raydir}{\mathbf{d}}

\newcommand{\viewdir}{\mathbf{d}}

\newcommand{\feat}{\mathbf{f}}

\newcommand{\mipmap}{\mathcal{R}}
\newcommand{\plane}{\mathcal{P}}
\newcommand{\density}{\tau}
\newcommand{\col}{{c}}

\newcommand{\baseradius}{\dot r}

\newcommand{\vs}{\textit{vs.}}

\def\etc{etc.\ }
\newcommand{\eg}{\textit{e.g.}}
\newcommand{\ie}{\textit{i.e.}}

\newcommand{\aka}{\textit{a.k.a.}}

\newcommand{\zval}{t}

\newcommand{\transpose}{{\operatorname{T}}}

\makeatletter
\@ifundefined{And}{\def\And{\&}}{}
\makeatother
\begin{document}
% Title portion
\title[Rip-NeRF: Anti-aliasing Radiance Fields with Ripmap-Encoded Platonic Solids]{Rip-NeRF: Anti-aliasing Radiance Fields
with Ripmap-Encoded Platonic Solids}

% DO NOT ENTER AUTHOR INFORMATION FOR ANONYMOUS TECHNICAL PAPER SUBMISSIONS TO SIGGRAPH 2024!
\author{Junchen Liu}
\authornote{Equal Contribution}
\orcid{0009-0003-4692-7776}
\affiliation{
 \institution{Beihang University}
 \city{Beijing}
 \country{China}}
\email{liujunchen0214@gmail.com}

\author{Wenbo Hu}
\authornotemark[1]
\orcid{0000-0001-6082-4966}
\affiliation{
 \institution{Tencent AI Lab}
 \city{Shenzhen}
 \country{China}}
\email{wbhu@tencent.com}

\author{Zhuo Yang}
\authornotemark[1]
\orcid{0009-0001-5572-1679}
\affiliation{
 \institution{Beijing Institute of Technology}
 \city{Beijing}
 \country{China}}
\email{zhuoyang@bit.edu.cn}

\author{Jianteng Chen}
\orcid{0009-0002-5442-088X}
\affiliation{
 \institution{Beijing Institute of Technology}
 \city{Beijing}
 \country{China}}
\email{chenjiantengx@gmail.com}

\author{Guoliang Wang}
\orcid{0009-0009-2855-8990}
\affiliation{
 \institution{Tsinghua University}
 \city{Beijing}
 \country{China}}
\email{wanggl199705@gmail.com}

\author{Xiaoxue Chen}
\orcid{0000-0001-6587-6060}
\affiliation{
 \institution{Tsinghua University}
 \city{Beijing}
 \country{China}}
\email{chenxx21@mails.tsinghua.edu.cn}

\author{Yantong Cai}
\orcid{0000-0003-2137-4979}
\affiliation{
 \institution{Dermatology Hospital}
 \city{Guangzhou}
 \country{China}}
\affiliation{
 \institution{Southern Medical University}
 \city{Guangzhou}
 \country{China}}
\email{yangtcai@gmail.com}

\author{Huan-ang Gao}
\orcid{0009-0004-6727-5778}
\affiliation{
 \institution{Tsinghua University}
 \city{Beijing}
 \country{China}}
\email{gha20@mails.tsinghua.edu.cn}

\author{Hao Zhao}
\authornote{Corresponding Author}
\orcid{0000-0001-7903-581X}
\affiliation{
 \institution{Tsinghua University}
 \city{Beijing}
 \country{China}}
\email{zhaohao@air.tsinghua.edu.cn}

%\authorsaddresses{}

\renewcommand\shortauthors{Liu, Hu, Yang et al.}

\begin{abstract}
    Despite significant advancements in Neural Radiance Fields (NeRFs), the renderings may still suffer from aliasing and blurring artifacts, since it remains a fundamental challenge to effectively and efficiently characterize \emph{anisotropic areas} induced by the cone-casting procedure.
    This paper introduces a \emph{Ripmap-Encoded Platonic Solid} representation to precisely and efficiently featurize 3D \emph{anisotropic} areas, achieving high-fidelity anti-aliased renderings.
    Central to our approach are two key components: \emph{Platonic Solid Projection} and \emph{Ripmap encoding}.
    The Platonic Solid Projection factorizes the 3D space onto the unparalleled faces of a certain Platonic solid, such that the anisotropic 3D areas can be projected onto planes with distinguishable characterization.
    Meanwhile, each face of the Platonic solid is encoded by the Ripmap encoding, which is constructed by anisotropically pre-filtering a learnable feature grid, to enable featurzing the projected anisotropic areas both precisely and efficiently by the anisotropic area-sampling.
    Extensive experiments on both well-established synthetic datasets and a newly captured real-world dataset demonstrate that our Rip-NeRF attains state-of-the-art rendering quality, particularly excelling in the fine details of repetitive structures and textures, while maintaining relatively swift training times, as shown in Fig.\ref{fig:teaser}.
    The source code and data for this paper are at \url{https://github.com/JunchenLiu77/Rip-NeRF}.
\end{abstract}

\begin{teaserfigure}
    \centering
    \vspace{-4mm}
    \includegraphics[width=1.0\linewidth]{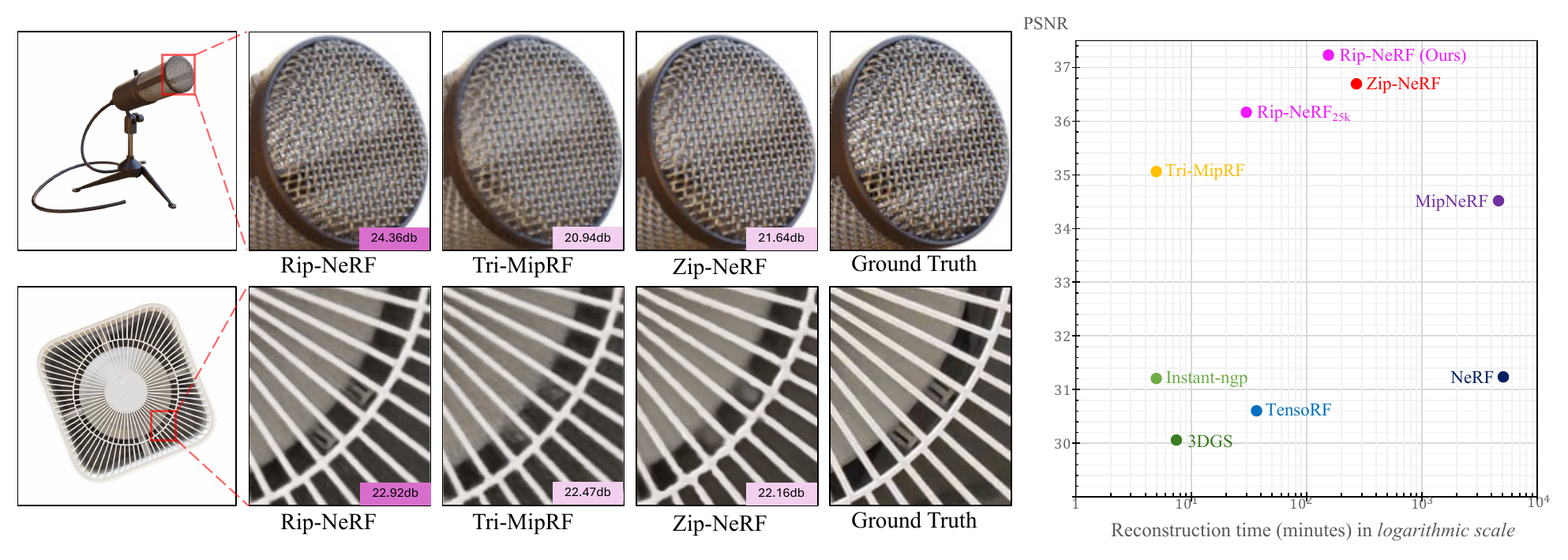}
    \vspace{-8mm}
    \caption{
        Qualitative and quantitative results of our Rip-NeRF and  several representative baseline methods, \eg, Zip-NeRF~\cite{barron2023zipnerf}, Tri-MipRF~\cite{hu2023Tri-MipRF}, \etc
        Rip-NeRF$_\text{25k}$ is a variant of Rip-NeRF that reduces the training iterations from $120 k$ to $25 k$ for better efficiency.
        The first and second rows in the left panel are results from the multi-scale Blender dataset~\cite{barron2021mip} and our newly captured real-world dataset, respectively.
        Our Rip-NeRF can render high-fidelity and anti-aliasing images from novel viewpoints while maintaining efficiency.
    }
    \Description{Qualitative and quantitative results of our Rip-NeRF and several representative baseline methods, including Zip-NeRF, Tri-MipRF, etc. Rip-NeRF 25k is a variant of Rip-NeRF that reduces the training iterations from 120k to 25k for better efficiency. Our Rip-NeRF can render high-fidelity and anti-aliasing images from novel viewpoints while maintaining efficiency.}
    % \vspace{-1em}
    \label{fig:teaser}
\end{teaserfigure}





\maketitle

\section{Introduction}
\label{sec:intro}

% NeRF
The epoch-making Neural Radiance Fields (NeRFs)~\cite{mildenhall2020nerf} employ a neural network to represent the scene as a continuous 5D function, which is defined as the radiance and density along a ray at a certain 3D location and direction.
They have promoted significant progress in numerous tasks, \eg, novel view synthesis~\cite{chen2022tensorf,martin2021nerf,yuan2023slimmerf}, geometry reconstruction~\cite{wang2021neus,chen2023nerrf}, content generation~\cite{chan2022efficient,peng2023synctalk}, simulation~\cite{wu2023mars,wei2024editable} and automation~\cite{zhu2022nice,park2023camp,zhu2023latitude,zhou2024pad}.

\begin{figure}[!t]
	\centering
	\includegraphics[width=1.0\linewidth]{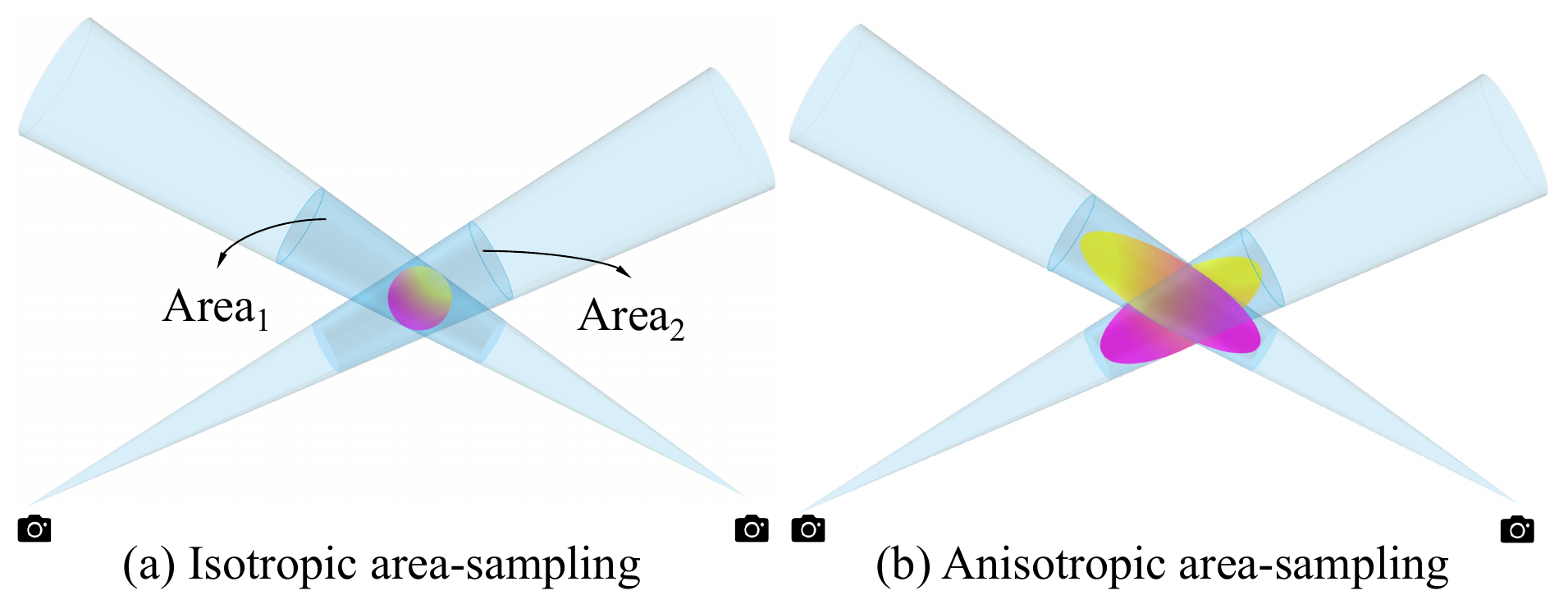}
        \Description{Difference between isotropic area-sampling and anisotropic area-sampling. The two anisotropic areas from different cones are ambiguously mapped to the same sampling area under the isotropic area-sampling, while are distinguishable under anisotropic area-sampling.}
	\vspace{-2em}
	\caption{
		The two anisotropic areas $\text{Area}_1$ and $\text{Area}_2$ from different cones are ambiguously mapped to the same sampling area under the isotropic area-sampling (a), while are distinguishable under anisotropic area-sampling (b).}
	\vspace{-2em}
	\label{fig:cone_aniso}
\end{figure}

% NeRF -> Mip-NeRF
The aliasing artifacts remain a challenging problem in NeRFs, which are caused by the discrete sampling of the continuous physical world.
As a pioneer, Mip-NeRF~\cite{barron2021mip} proposed an analytical integrated positional encoding for the purely implicit representation to facilitate anti-aliasing rendering.
However, both the training speed and rendering quality of it are limited by the purely implicit representation.
To this end, Zip-NeRF~\cite{barron2023zipnerf} and Tri-MipRF~\cite{hu2023Tri-MipRF} proposed anti-aliasing mechanisms based on the hybrid representation in the form of multi-sampling and area-sampling (\aka pre-filtering), respectively.
Nevertheless, multi-sampling inherently demands a large number of samples to featurize a single area, which puts it in a dilemma between the rendering quality and computational overhead.
On the other hand,
area-sampling of Tri-MipRF is more efficient as it can directly featurize a sub-volume. However, its isotropic mechanism significantly limits the ability to represent anisotropic areas that are ubiquitously induced by cone casting methods~\cite{barron2022mip, hu2023Tri-MipRF} in the volume rendering.
As shown in Fig.~\ref{fig:cone_aniso} (a), the isotropic area-sampling cannot differentiate the anisotropic areas from different cones, which leads to ambiguities in the representation.
Consequently, as shown in Fig.~\ref{fig:teaser}, blurriness on the surface of the microphone can be observed under this isotropic area-sampling.

% Tri-MipRF -> Rip-NeRF
In this paper, we propose a Ripmap-encoded Platonic solids representation, termed \emph{Rip-NeRF}, for high-fidelity anti-aliased neural radiance fields.
It enables featurizing 3D areas more precisely with various shapes using only one sample per area, such that anisotropic areas from different cones are distinguishable under our representation, as shown in Fig.~\ref{fig:cone_aniso} (b).
The key to achieving this lies in two techniques, \ie, the
\emph{Platonic Solid Projection}
and the \emph{Ripmap Encoding}.
On one hand, the Platonic Solid Projection is a 3D space factorization method, which projects 3D areas onto the unparalleled faces of Platonic solids.
By doing so, we can precisely represent 3D scenes with 2D feature grids, rather than 3D volumes, and the memory consumption is significantly reduced from $O(n^3)$ to $O(n^2)$.
Note that, the orthogonal tri-plane representation adopted in~\cite{chan2022efficient, hu2023Tri-MipRF} can be derived from a regular hexahedron (cube) in this perspective.
On the other hand, the Ripmap Encoding can featurize the projected anisotropic 2D areas with a \emph{Ripmap}~\cite{mcreynolds1998advgrphics} (\aka anisotropic Mipmap), which is a feature grid pre-filtered with anisotropic kernels to represent the face of the Platonic solid.
Compared with the tri-plane factorization and isotropic area-sampling (Mipmap) of Tri-MipRF~\cite{hu2023Tri-MipRF}, our Platonic Solid Projection together with the Ripmap Encoding enables more precisely featurizing anisotropic 3D areas in an efficient pre-filtering manner, such that sharper and more accurate details on the repetitive patterns of microphone can be reconstructed and rendered as shown in Fig.~\ref{fig:teaser}.

%-----------------------------------------------------------------------------
%-----------------------------------------------------------------------------

To evaluate the effectiveness of our Rip-NeRF, we conduct extensive experiments on both well-established public benchmarks and a newly captured real-world dataset, where the quantitative and qualitative results reveal that our Rip-NeRF achieves state-of-the-art rendering quality while maintaining efficient reconstruction.
Besides, the ablation studies also demonstrate the effectiveness of our individual proposed components, \ie the Platonic Solid Projection and the Ripmap Encoding.
Furthermore, our Platonic Solid Projection introduces a flexible trade-off between rendering quality and efficiency, \eg, training time and GPU memory consumption, by selecting different Platonic solids with a certain number of faces.
% summary
Our contributions are summarized below.
\begin{itemize}
	\item We propose a 3D space factorization method, \emph{Platonic Solid Projection}, to represent a 3D scene with the 2D faces of a Platonic solid, such that the anisotropic 3D areas can be projected onto planes with distinguishable characterization.
 
	\item We propose to represent the faces of a Platonic solid by \emph{Ripmap Encoding}, such that the projected anisotropic 2D areas can be precisely and efficiently featurized by the anisotropic area-sampling.

	\item Our \emph{Rip-NeRF} achieves state-of-the-art rendering quality on both the public benchmarks and a newly captured real-world dataset while maintaining efficient reconstruction. And it enables a flexible trade-off between quality and efficiency.
\end{itemize}
% %%%%%%%%%%%%%%%%%%%%%%%%%%%%%%%%%%%%%%%%%%%%%%%%
\begin{figure*}[!t]
    \centering
    \includegraphics[width=1.0\linewidth]{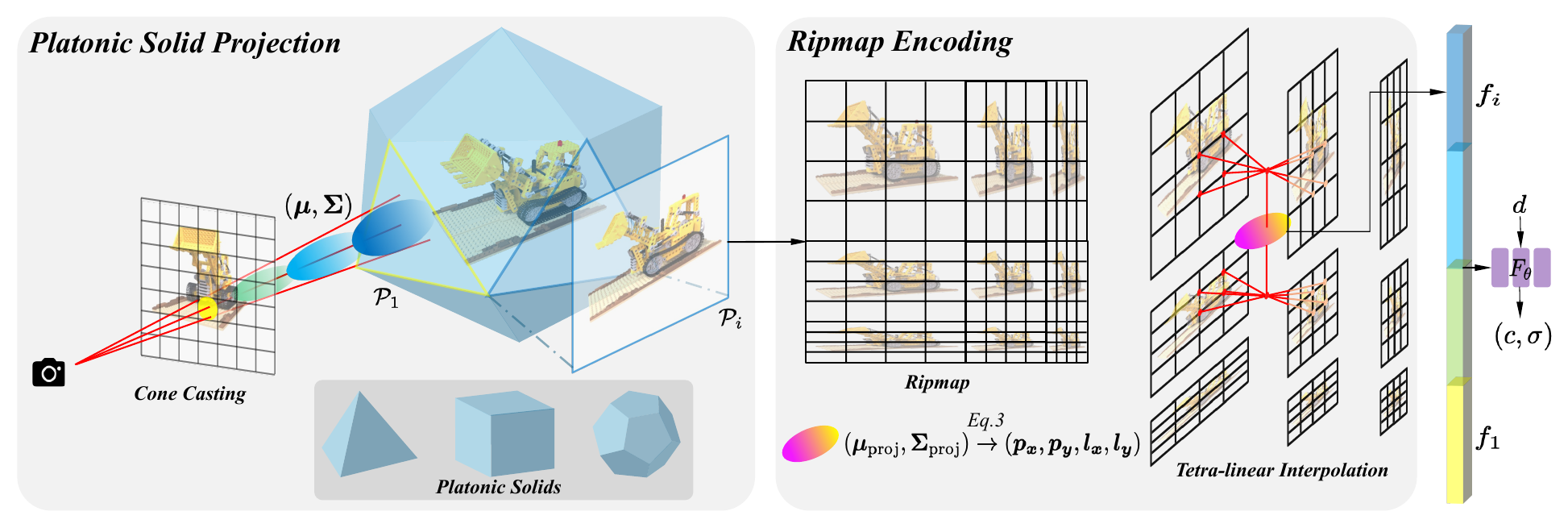}
    \Description{Difference between isotropic area-sampling and anisotropic area-sampling. The two anisotropic areas from different cones are ambiguously mapped to the same sampling area under the isotropic area-sampling, while are distinguishable under anisotropic area-sampling.}
    \vspace{-8mm}
    \caption{
        Overview of our Rip-NeRF. We first cast a cone for each pixel, and then divide the cone into multiple conical frustums, which are further characterized by anisotropic 3D Gaussians parameterized by their mean and covariance $(\boldsymbol{\mu}, \; \boldsymbol{\Sigma})$.
        Next, to featurize a 3D Gaussian, we project it onto the unparalleled faces of the Platonic solid, denoted as $\{\mathcal{P}_i \; | \; i=1,...,n\}$ to form a 2D Gaussian $(\boldsymbol{\mu}_\text{proj}, \; \boldsymbol{\Sigma}_\text{proj})$, while the Platonic solid's faces are represented by the Ripmap Encoding with learnable parameters.
        Subsequently, we perform tetra-linear interpolation on the Ripmap Encoding to query corresponding feature vectors $f_i$ for the 2D Gaussian, where the position $(p_x, \; p_y)$ and level $(l_x, \; l_y)$ used in the interpolation are determined by the mean and covariance $(\boldsymbol{\mu}_\text{proj}, \; \boldsymbol{\Sigma}_\text{proj})$ of the 2D Gaussian, respectively.
        Finally, feature vectors $f_i$ from all Platonic solids' faces and the encoded view direction $d$ are aggregated together to estimate the color $c$ and density $\sigma$ of the conical frustums by a tiny MLP $F_\theta$.
    }
    \vspace{-4mm}
    \label{fig:overview}
\end{figure*}
% %%%%%%%%%%%%%%%%%%%%%%%%%%%%%%%%%%%%%%%%%%%%%%%%

\section{Related Work}

\subsection{Hybrid Representations in Neural Rendering}
With the rise of deep learning, neural rendering, especially neural radiance fields (NeRF)~\cite{mildenhall2020nerf}, has drawn increasing attention in 
novel view synthesis~\cite{Nerfstudio, Variablebitrateneuralfields, Eikonalfields, Relufields, CamP, Differentiable-point-based-radiance-fields, Discontinuity-Aware, NeuralCaches}, 
avatar reconstruction~\cite{LatentAvatar, Morf, BakedAvatar, NeRSemble, HumanRF, AvatarReX, SAILOR, NeRFFaceLighting, SketchFaceNeRF, Live3DPortrait}, 
reconstruction from sparse views~\cite{ViP-NeRF, SimpleNeRF, Fdnerf, lao2024corresnerf},
reconstruction of large-scale scenes~\cite{ScaNeRF},
scene editing~\cite{SeamlessNeRF, NeRF-texture, Dreameditor, RelightingNeuralRadianceFields, De-nerf, ActRay,  NeRFFaceEditing}, 
and dynamic scene reconstruction~\cite{Efficient-neural-radiance, Hypernerf}.
NeRF methods typically rely on neural networks to act as continuous representations. 
However, the pure implicit representations are computationally intensive and hard to represent high-frequency details.

On the other hand, recent works explored representing 3D scenes with explicit data structures, \eg, octrees~\cite{yu2021plenoctrees, RT-Octree}, sparse voxels~\cite{fridovich2022plenoxels}, and VDB~\cite{hyan2023plenvdb}.
Nevertheless, explicit representations often suffer from large storage footprints and low rendering quality.
To this end, hybrid representations, combining a tiny MLP and explicit data structures, like hash table~\cite{muller2022instant}, tri-plane~\cite{chan2022efficient}, and Vector-Matrix~\cite{chen2022tensorf}, have emerged to improve rendering quality and efficiency~\cite{Merf, smerf, MCNeRF, Dictionaryfields}.
But these representations still suffer from aliasing artifacts, due to the ray casting procedure that discretely samples a continuous signal.

\subsection{Anti-Aliasing in Neural Radiance Fields}

Essentially, aliasing occurs as the overlapping frequency components due to insufficient sampling rates. 
Therefore, to alleviate this issue, we can directly increase the sampling rate by multi-sampling, or appropriately decrease the frequency of the scene by pre-filtering (\aka area-sampling).
In the neural radiance fields (NeRF)~\cite{mildenhall2020nerf} context, Mip-NeRF~\cite{barron2021mip} pioneered the anti-aliasing for NeRF by the integrated positional encoding that enables area-sampling.
Mip-NeRF 360~\cite{barron2022mip} further explored the anti-aliasing for unbounded scenes to improve the applicability.
However, both the training and rendering of them are computationally intensive due to their pure implicit representation.

Recently, Zip-NeRF~\cite{barron2023zipnerf} presented a multi-sampling strategy to enable anti-aliasing for more efficient hybrid grid-based representation~\cite{muller2022instant}.
However, multi-sampling inherently demands many samples to featurize a single area, which puts it in a dilemma between the rendering quality and computational overhead.
Conversely, Tri-MipRF~\cite{hu2023Tri-MipRF} proposed an area-sampling strategy that models the scene as three orthogonal mipmaps, benefiting from the efficiency and compactness of the hybrid plane-based representation~\cite{chan2022efficient}.
Nevertheless, its isotropic mechanism significantly limits the ability to represent anisotropic areas induced by cone casting.
In contrast, our method not only enables precisely featurizing anisotropic 3D areas in an efficient area-sampling manner but also maintains the compactness of the hybrid plane-based representation.
\section{Method}

\subsection{Overview}
Given a set of calibrated multi-view images, our goal is to render \emph{high-fidelity anti-aliasing} images in a NeRF~\cite{mildenhall2020nerf} fashion.
Following~\cite{muller2022instant,barron2023zipnerf,hu2023Tri-MipRF,chen2022tensorf}, our Rip-NeRF adopts a hybrid representation, to benefit from both the efficiency and flexibility of explicit and implicit representations.
As shown in Fig.~\ref{fig:overview}, to render a pixel, we cast a cone for it and divide the cone into multiple conical frustums, similar to~\cite{barron2021mip,barron2022mip,barron2023zipnerf,hu2023Tri-MipRF}, which effectively avoids the discrete point-based sampling of the continuous signals in the image plane.
To featurize a conical frustum, multi-sampling~\cite{barron2023zipnerf} and area-sampling~\cite{hu2023Tri-MipRF} are two types of strategies in the hybrid representation.
For the efficiency consideration, we adopt the latter one to first characterize the conical frustum by an anisotropic 3D Gaussian~\cite{barron2021mip} with the mean and covariance as $(\boldsymbol{\mu}, \boldsymbol{\Sigma})$, and then featurize it with our proposed \emph{Platonic Solid Projection} and \emph{Ripmap Encoding}.
Different from the Tri-Mip encoding~\cite{hu2023Tri-MipRF} which roughly characterizes the frustum as isotropic balls, our Platonic Solid Projection together with Ripmap Encoding can accurately featurize the anisotropic 3D Gaussians, which are to be presented in the following sections.
After featurizing the conical frustums, we employ a tiny MLP $F_\theta$ to estimate the color $c$ and density $\sigma$ of the frustums, and then render the pixel color by the volume rendering~\cite{mildenhall2020nerf}.
The whole system is optimized end-to-end with a photometric loss between the rendered and observed images.
% %%%%%%%%%%%%%%%%%%%%%%%%%%%%%%%%%%%%%%%%%%%%%%%%
\begin{figure}[!t]
    \centering
    \includegraphics[width=0.9\linewidth]{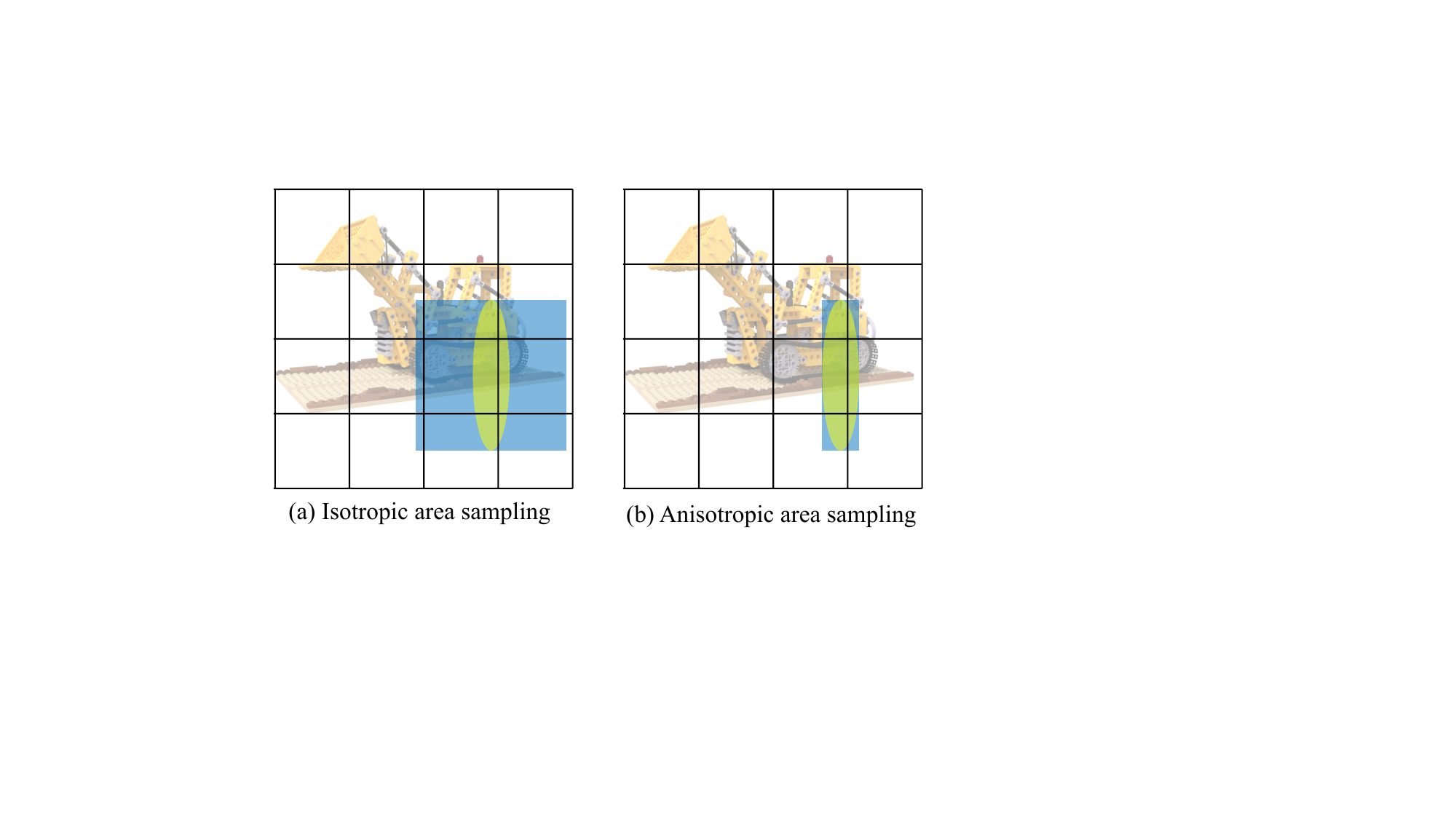}
    \Description{Comparison between isotropic and anisotropic area sampling in Mipmap and Ripmap, for characterizing the projected Gaussian.}
    \vspace{-4mm}
    \caption{
        Comparison between isotropic (a) and anisotropic (b) area sampling in Mipmap and Ripmap, for characterizing the projected Gaussian.
    }
    \vspace{-6mm}
    \label{fig:demo_aniso}
\end{figure}
% %%%%%%%%%%%%%%%%%%%%%%%%%%%%%%%%%%%%%%%%%%%%%%%%

\subsection{Ripmap Encoding}
Before introducing how to featurize an anisotropic 3D Gaussian by projection, we first present the featurization of an anisotropic 2D Gaussian, which is the basis of the 3D case.
Tri-MipRF~\cite{hu2023Tri-MipRF} proposed to use a 2D Mipmap containing learnable features to support area sampling of an \emph{isotropic} disc.
However, as shown in Fig.~\ref{fig:demo_aniso} (a), the isotropic Mipmap structure of Tri-MipRF makes its sampling area a square, which can not precisely characterize the projected anisotropic 2D Gaussian, whose axis-aligned bounding-box is essentially a rectangle.
In conventional graphics, anisotropic area sampling is proposed to address the aliasing issue when the view direction closely aligns with an axis of the UV texture since the occurrence of aliasing is pronounced due to the significant variance in sampling density across the texture space.
Among various solutions~\cite{4056764}, Ripmap~\cite{mcreynolds1998advgrphics} is a popular one due to its effectiveness and simplicity, which dynamically adjusts the
the aspect ratio of the pre-filtering kernel based on the angle of incidence.
Inspired by this, we propose \emph{Ripmap Encoding} to employ a Ripmap with learnable parameters to enable anisotropic area-sampling in the neural rendering context, such that the anisotropic ellipsoidal footprint of a Gaussian can be characterized more precisely, as shown in Fig.~\ref{fig:demo_aniso} (b).

\paragraph{Ripmap Encoding construction.}
The Ripmap Encoding $\mipmap$ contains $L \times L$ levels, while the base level $\mipmap^{0, \; 0}$ is a 2D feature grid $\mathcal{F}$ with the shape of $H \times W \times C$, where the $H, W, C$ are the height, width, and number of channels, respectively.
Other levels are constructed by performing anisotropic average pooling $\boldsymbol{\text{Avg}_{2 \times 1}}$ and $\boldsymbol{\text{Avg}_{1 \times 2}}$ on the lower level feature grid:
\begin{equation}
    \begin{aligned}
        \label{eq:construct_animip}
        \mipmap^{i, \; j} = \left\{
        \begin{array}{ll}
            \boldsymbol{\text{Avg}_{2 \times 1}}\big(\mipmap^{i, \; j-1}\big) & \text{if } j \neq 0            \\
            \boldsymbol{\text{Avg}_{1x2}}\big(\mipmap^{i-1, \; j}\big)        & \text{if } i \neq 0 \And j = 0 \\
            \mathcal{F}                                                & \text{otherwise,}
        \end{array}
        \right.
    \end{aligned}
\end{equation}
where $i$ and $j$ are the indices of the levels in the $x$ and $y$ directions, respectively.
Note that, only the base level $\mipmap^{0, \; 0}$ is learnable, while other levels are derived from it, which makes the Ripmap Encoding compact and consistent among levels.

\paragraph{Ripmap Encoding querying.}
Once the Ripmap Encoding is constructed, we can featurize an anisotropic 2D Gaussian by querying the Ripmap Encoding using the tetra-linear interpolation:

\begin{equation}
    \begin{aligned}
        \label{eq:featurization}
        \boldsymbol{f} & = \mipmap(p_x,\; p_y, \; l_x, \; l_y),
    \end{aligned}
\end{equation}
where $(p_x, \; p_y)$ and $(l_x, \; l_y)$ are the position and level used in the interpolation, respectively. The formal mathematical expression of tetra-linear interpolation is presented in the supplementary material.
Since the querying position and level correspond to the location and size of the sampling area, respectively, we derive them from the mean and covariance $(\boldsymbol{{\mu_\text{proj}}}, \; \boldsymbol{{\Sigma_\text{proj}}})$ of the Gaussian as:
\begin{equation}
    \begin{aligned}
        \label{eq:compute_level}
        p_d & = \mu_d                                                        \\
        l_d & = \log_2\bigg(\frac{w\sigma_d}{r}\bigg), \quad d \in \{x, y\},
    \end{aligned}
\end{equation}
where $\sigma_x, \; \sigma_y = \sqrt{\text{diag}(\boldsymbol{\Sigma_\text{proj}})}$ are the standard deviations along the $\textbf{x}$ and $\textbf{y}$ axes, respectively, $w$ is a hyper-parameter to adjust how much probability mass of the Gaussian footprint is covered, and $r$ is the radius of the bounding sphere for the reconstructed scene.

 %%%%%%%%%%%%%%%%%%%%%%%%%%%%%%%%%%%%%%%%%%%%%%%%
\begin{figure}[!t]
    \centering
    \includegraphics[width=1.0\linewidth]{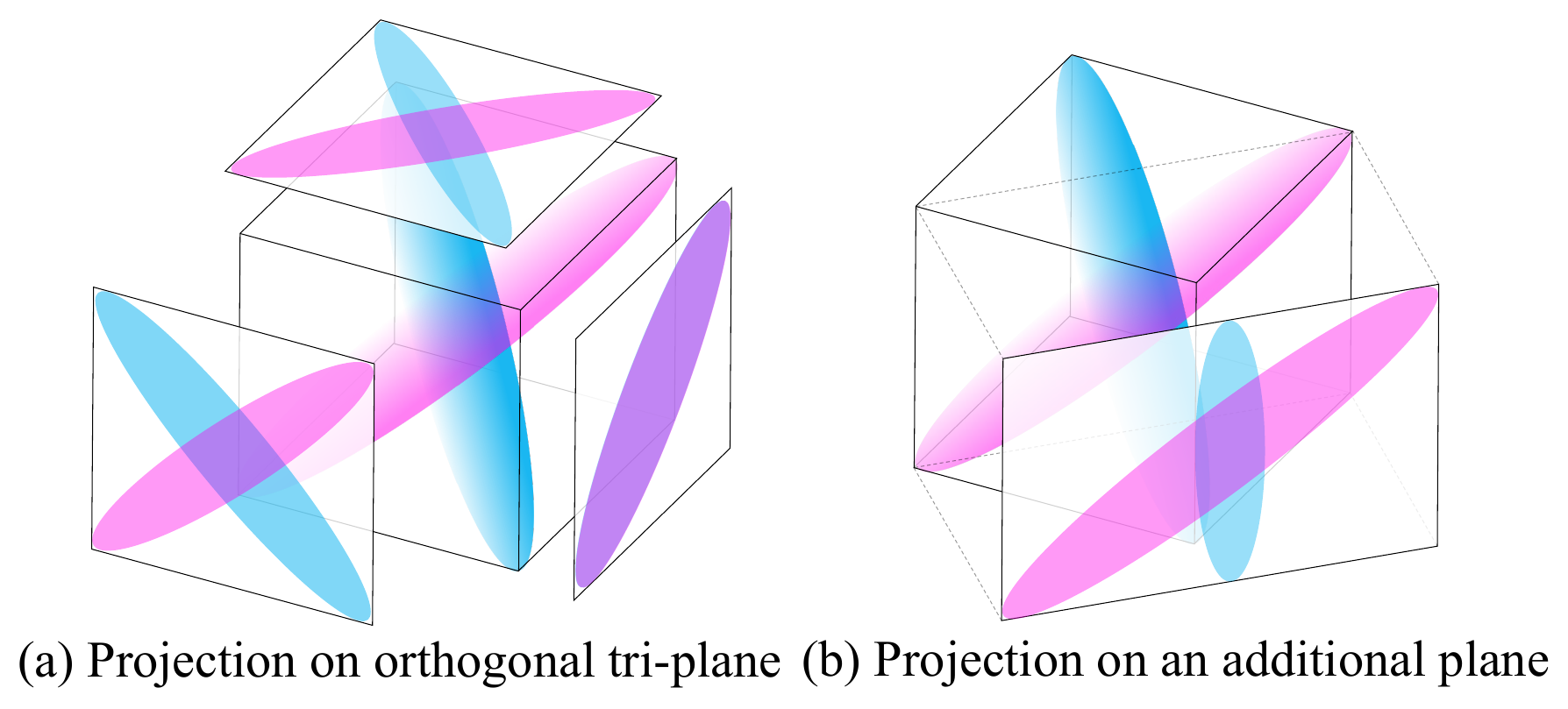}
    \Description{The necessity of using multi-planes. Two 3D ellipsoids, whose major axes are aligned along two different body diagonals of a cube, share the same 2D AABBs on the orthogonal tri-plane, making them indistinguishable under the Ripmap encoding. However, their difference can be captured by an additional different-oriented plane.}
    \vspace{-6mm}
    \caption{
        Two 3D ellipsoids, whose major axes are aligned along two different body diagonals of a cube, share the same 2D AABBs on the orthogonal tri-plane (a), making them indistinguishable under the Ripmap encoding.
        However, their difference can be captured by an additional different-oriented plane (b).
    }
    \vspace{-4mm}
    \label{fig:demo_multi}
\end{figure}
% %%%%%%%%%%%%%%%%%%%%%%%%%%%%%%%%%%%%%%%%%%%%%%%%

\begin{table*}[]
    \renewcommand{\tabcolsep}{1pt}
    % \vspace*{0.5em}
    \centering
    \resizebox{\linewidth}{!}{
        \begin{tabular}{@{}l@{\,\,}|c@{\,}|c@{\,}|ccccp{0.9cm}<{\centering}|ccccp{0.9cm}<{\centering}|ccccp{0.9cm}<{\centering}}
             &                    &                   & \multicolumn{5}{c|}{PSNR $\uparrow$} & \multicolumn{5}{c|}{SSIM $\uparrow$} & \multicolumn{5}{c}{LPIPS $\downarrow$}                                                                                                                                                                                                                            \\
             & Train $\downarrow$ & Size $\downarrow$ & Full Res.                            & $\nicefrac{1}{2}$ Res.               & $\nicefrac{1}{4}$ Res.                 & $\nicefrac{1}{8}$ Res. & Avg. & Full Res. & $\nicefrac{1}{2}$ Res. & $\nicefrac{1}{4}$ Res. & $\nicefrac{1}{8}$ Res. & Avg. & Full Res. & $\nicefrac{1}{2}$ Res. & $\nicefrac{1}{4}$ Res. & $\nicefrac{1}{8}$ Res & Avg. \\ \hline
            NeRF w/o $\mathcal{L}_\text{area}$ & 3 days & \cellcolor{orange}5.00 MB &
31.20 & 30.65 & 26.25 & 22.53 &  27.66 & 
0.950 & 0.956 & 0.930 & 0.871 & 0.927 & 
0.055 & 0.034 & 0.043 & 0.075 & 0.052
\\
NeRF~\cite{mildenhall2020nerf} & 3 days & \cellcolor{orange}5.00 MB &
29.90 & 32.13 & 33.40 & 29.47 &  31.23 & 
0.938 & 0.959 & 0.973 & 0.962 &  0.958 & 
0.074 & 0.040 &  0.024 & 0.039 & 0.044 
\\
TensoRF~\cite{chen2022tensorf} & 19 mins & 71.8 MB &
32.11 & 33.03 & 30.45 &  26.80 & 30.60 & 
0.956 & 0.966 & 0.962 & 0.939 & 0.956 & 
0.056 & 0.038 & 0.047 & 0.076 & 0.054 
\\
Instant-NGP~\cite{muller2022instant} & \cellcolor{red}5 mins & 64.1 MB &
30.00 & 32.15 & 33.31 & 29.35 & 31.20 & 
0.939 & 0.961 & 0.974 & 0.963 & 0.959 & 
0.079 & 0.043 & 0.026 & 0.040 & 0.047  
\\
Mip-NeRF~\cite{barron2021mip}   & 3 days & \cellcolor{red}2.50 MB &
32.63 & 34.34 & 35.47 & 35.60 & 34.51 & 
0.958 & 0.970 & 0.979 & 0.983 & 0.973 & 
0.047 & \cellcolor{yellow}0.026 & 0.017 & \cellcolor{yellow}0.012 & 0.026 
\\
% 3DGS~\cite{kerbl3Dgaussians} & \cellcolor{yellow}7.5 min & \cellcolor{yellow}27 MB &
% 29.00 & 30.94 & 32.06 & 28.21 & 30.05 & 
% 0.946 & 0.965 & 0.976 & 0.964 & 0.963 & 
% 0.064 & 0.037 & 0.024 & 0.030 & 0.039
% \\

\hline
% Plenoxels~\cite{fridovich2022plenoxels} & 9 min & 778 MB &
% 31.60 & 32.85 & 30.26 & 26.63 & 30.34 & 
% 0.956 & 0.967 & 0.961 & 0.936 & 0.955 & 
% 0.052 & 0.032 & 0.045 & 0.077 & 0.051
% \\
% MipNeRF~\cite{barron2021mip}   & 3 days & \cellcolor{red}2.50 MB &
% 32.63 & 34.34 & 35.47 & 35.60 & 34.51 & 
% 0.958 & 0.970 & 0.979 & 0.983 & 0.973 & 
% 0.047 & \cellcolor{yellow}0.026 & 0.017 & 0.012 & 0.026 
% \\
Tri-MipRF~\cite{hu2023Tri-MipRF} & \cellcolor{orange}5.5 mins & 48.0 MB &
33.57 & 35.21 & 35.96 & 36.46 & 35.30 &
0.962 & 0.975 & 0.982 & 0.987 & 0.976 &
0.052 & 0.029 & 0.019 & 0.013 & 0.028
\\
Zip-NeRF~\cite{barron2023zipnerf} & 4.5 hrs & 592 MB &
\cellcolor{yellow}34.21 & \cellcolor{orange}36.55 & \cellcolor{orange}37.88 & \cellcolor{orange}38.13 & \cellcolor{orange}36.69 & 
\cellcolor{red}0.974 & \cellcolor{red}0.985 & \cellcolor{red}0.990 & \cellcolor{red}0.992 & \cellcolor{red}0.985 
 & \cellcolor{red}0.036 & \cellcolor{red}0.019 & \cellcolor{orange}0.014 & 0.015 & \cellcolor{orange}0.021 
\\
3DGS~\cite{kerbl3Dgaussians} & \cellcolor{yellow}7.5 mins & \cellcolor{yellow}27.0 MB &
29.00 & 30.94 & 32.06 & 28.21 & 30.05 & 
0.946 & 0.965 & 0.976 & 0.964 & 0.963 & 
0.064 & 0.037 & 0.024 & 0.030 & 0.039
\\

\hline
Rip-NeRF$_\text{25k}$ & 32 mins & 160 MB &
\cellcolor{orange}34.30 & \cellcolor{yellow}35.94 & \cellcolor{yellow}36.92 & \cellcolor{yellow}37.47 & \cellcolor{yellow}36.16 &
\cellcolor{yellow}0.966 & \cellcolor{yellow}0.978 & \cellcolor{yellow}0.984 & \cellcolor{yellow}0.989 & \cellcolor{yellow}0.979 & 
\cellcolor{yellow}0.045 & \cellcolor{orange}0.025 & \cellcolor{yellow}0.016 & \cellcolor{orange}0.011 & \cellcolor{yellow}0.024
\\
Rip-NeRF (Ours) & 2.6 hrs & 160 MB 
& \cellcolor{red}35.30 & \cellcolor{red}37.01 & \cellcolor{red}38.07 & \cellcolor{red}38.54 & \cellcolor{red}37.23 &
\cellcolor{orange}0.973 & \cellcolor{orange}0.983 & \cellcolor{orange}0.988 & \cellcolor{orange}0.991 & \cellcolor{orange}0.984 &
\cellcolor{orange}0.037 & \cellcolor{red}0.019 & \cellcolor{red}0.011 & \cellcolor{red}0.008 & \cellcolor{red}0.019
        \end{tabular}
    }
    \vspace{0.5em}
    \caption{
        Quantitative performance on the multi-scale Blender dataset~\cite{barron2021mip}. We compared our Rip-NeRF, and its variant, Rip-NeRF$_\text{25k}$, which reduces the training iterations from $120k$ to $25k$ for fast reconstruction, against several representative methods. The best, second-best, and third-best results are marked in red, orange, and yellow, respectively.
    }
    \vspace{-2em}
    \label{tab:avg_multiblender_results}
\end{table*}
% %%%%%%%%%%%%%%%%%%%%%%%%%%

\subsection{Platonic Solid Projection}

Factorizing 3D space into a group of 2D planes has been proven to be effective and compact~\cite{peng2020convolutional,chan2022efficient,hu2023Tri-MipRF,fridovich2023k,cao2023hexplane,chen2022tensorf}.
Therefore, with the presented Ripmap Encoding for 2D area-sampling, the key to precisely featurizing anisotropic 3D Gaussians is how to project them onto the 2D planes.
Tri-MipRF~\cite{hu2023Tri-MipRF} projects isotropic 3D spheres onto three orthogonal 2D planes, however, this strategy falls short when dealing with anisotropic 3D Gaussians.
The 3D Gaussians, produced by the cone casting, are almost randomly distributed in the Euclidean space with various shapes due to the arbitrariness of the intrinsic and extrinsic camera parameters.
Thus, different 3D Gaussians may have the same 2D Axis-Aligned Bounding Box (AABB) when being projected onto the planes, \eg, two 3D ellipsoids with major axes aligned along two different body diagonals of a cube in Fig.~\ref{fig:demo_multi} (a), which makes them indistinguishable under the Ripmap Encoding.
To address this issue, we propose to project the 3D Gaussians onto a larger number of planes, which are appropriately distributed in the 3D space, and then concatenate the features queried from those planes together, such that the 2D AABBs of different Gaussians can be more distinguishable and the derived features of Gaussians more discriminative, as shown in Fig.~\ref{fig:demo_multi} (b).

\paragraph{Orientations and axes of the planes.}
Based on the above example, we intuitively think the more diverse the planes are oriented, the better their representation capability. To verify it, we tried three methods to evenly distribute the planes (the Platonic solids' faces, golden spiral~\cite{keinert2015spherical}, and spherical blue noise~\cite{wong2018spherical} ). We also tried one control group method to adopt spherical white noise to determine the planes, which are slightly less diverse. We find the evenly distributed group performs much better than the control group and the three even groups have similar performance. 
Therefore, for simplicity and good performance, we opted for the \emph{Platonic Solid Projection} that projects anisotropic 3D Gaussians into the unparalleled faces of a specific Platonic solid, \ie, tetrahedron, cube, octahedron, dodecahedron, and icosahedron, whose faces are congruent (identical in shape and size) regular polygons that have equivalent dihedral angles.
Note that, Mip-NeRF 360~\cite{barron2022mip}also adopted a similar strategy to derive off-axis integrated positional encoding features for 3D Gaussians. However, we propose the Platonic solid projection to explicitly project Gaussians onto planes for the area sampling using Ripmap encoding.
Our Platonic Solid Projection also provides a flexible trade-off between rendering quality and efficiency by selecting different Platonic solids with a certain number of faces, as to be demonstrated in Tab.~\ref{tab:multi_scale_ablation}.
Specifically, we denote the planes and their outward normals as $\{\mathcal{P}_i \; | \; i=1,...,n\}$ and $\{\boldsymbol{\phi}_i \in \mathbb{R}^3\; | \; i=1,...,n\}$, respectively, where $n$ is the number of faces of the selected Platonic solid.
And the local 2D axes $\textbf{x}_i \in \mathbb{R}^3$ and $\textbf{y}_i \in \mathbb{R}^3$ of the plane $\mathcal{P}_i$, which is used to allocate grids for the Ripmap Encoding, should be perpendicular to the plane's normal $\boldsymbol{\phi}_i$. Given the unit vectors $\boldsymbol{X}$, $\boldsymbol{Y}$ and $\boldsymbol{Z}$ of the world coordinate system, we empirically define them as:
\begin{equation}
    \begin{aligned}
        \label{eq:confim_axes}
        \begin{array}{lll}
            \textbf{x}_i = \boldsymbol{X} \quad                        & \textbf{y}_i = \boldsymbol{Y}

                                                                       & \text{if } \boldsymbol{\phi_i} = \boldsymbol{Z},                        \\
            \textbf{x}_i = \boldsymbol{Z} \times \boldsymbol{\phi}_i \quad & \textbf{y}_i = \textbf{x}_i \times \boldsymbol{\phi}_i & \text{otherwise }
        \end{array}
    \end{aligned}
\end{equation}
\paragraph{Featurization of anisotropic 3D Gaussians.}
After defining the orientations and axes of the planes, we can project the anisotropic 3D Gaussians, characterized by mean $\boldsymbol{\mu}$ and covariance $\boldsymbol{\Sigma}$, onto each plane $\plane_i$:
\begin{equation}
    \begin{aligned}
        \label{eq:project_gaussian}
        \mathcal{M}_i                     & = [\textbf{x}_i, \textbf{y}_i],
        \\
        \boldsymbol{\mu_\text{proj}^i}    & = \mathcal{M}_i^T \boldsymbol{\mu},
        \\
        \boldsymbol{\Sigma_\text{proj}^i} & = \mathcal{M}_i^T \boldsymbol{\Sigma} \mathcal{M}_i,
    \end{aligned}
\end{equation}
where $\mathcal{M}_i \in \mathbb{R}^{3 \times 2}$ is the projection matrix for mapping the 3D world coordinate system into the 2D coordinate system of plane $\plane_i$, $\boldsymbol{\mu_\text{proj}^i}$ and $\boldsymbol{\Sigma_\text{proj}^i}$ are the mean and covariance of the projected 2D Gaussians on plane $\plane_i$.
Then, we can query a feature vector $\boldsymbol{f}_i$ from the Ripmap Encoding of each plane $\plane_i$ using Eq.~\ref{eq:featurization} and Eq.~\ref{eq:compute_level}.
Finally, we concatenate all the feature vectors $\{\boldsymbol{f}_i \; | \; i=1,...,n\}$ from the corresponding planes $\{\plane_i \; | \; i=1,...,n\}$ to form the final feature vector $\boldsymbol{f}$ of the 3D Gaussian.

\section{Experiments}

\subsection{Implementation Details}
Since our Rip-NeRF acquires the 3D structure only from the calibrated multi-view 2D images, we optimize the whole system end-to-end with a photometric loss, assessing the discrepancy between rendered pixels and captured images.
To make the photometric loss area-aware, we scale the loss for each pixel by incorporating the area of its footprint on the image plane, denoted as ``area loss'' $\mathcal{L}_\text{area}$, following~\cite{barron2021mip,hu2023Tri-MipRF}.
Except for the ablation study, we set the shape of the base level in the Ripmap Encoding $\mipmap^{0, \; 0}$ to $H=512$, $W=512$, and $C=16$, aligned with Tri-MipRF~\cite{hu2023Tri-MipRF}.
For the Platonic Solid Projection, we by default employ the icosahedron, which contains ten unparalleled faces.
We set the value of hyper-parameter $w$ in \cref{eq:compute_level} to $2.0$, which is used to adjust how much probability mass of the Gaussian footprint is covered.

For efficiency, we not only employ the tiny-cuda-nn~\cite{tiny-cuda-nn} library for its highly optimized MLP implementation but also implement our Platonic Solid Projection and Ripmap Encoding in CUDA kernels. 
Besides, we also adopt the NerfAcc~\cite{li2022nerfacc} library to incorporate a binary occupancy grid to indicate empty \vs non-empty space similar to~\cite{muller2022instant,hu2023Tri-MipRF}, which enables efficiently skipping samples in the empty area.
All the modules are integrated into the PyTorch framework~\cite{paszke2019pytorch} since PyTorch is widely used in the research community.
Training of our Rip-NeRF is conducted using the AdamW optimizer~\cite{loshchilov2019decoupled} with $120k$ iterations.
We also present a variant of our method, Rip-NeRF$_\text{25k}$, which is trained with $25k$ iterations, for faster reconstruction that only slightly sacrifices the rendering quality.
We apply a weight decay of $1\times10^{-5}$ and an initial learning rate of $2\times10^{-3}$, which is modulated using PyTorch's MultiStepLR scheduler.
Following Tri-MipRF~\cite{hu2023Tri-MipRF}, the learning rate for Ripmap encoding is scaled up by $10.0$ times since it directly represents the scene.

\subsection{Evaluation on the Muti-scale Blender Dataset}

Following Mip-NeRF~\cite{barron2021mip} and Tri-MipRF~\cite{hu2023Tri-MipRF}, to evaluate the capability of rendering anti-aliasing and fine image details, we benchmarked our Rip-NeRF on the multi-scale Blender dataset~\cite{barron2021mip}, which is a combination of the Blender dataset~\cite{mildenhall2020nerf} and its down-scaled versions with a factor of 2, 4, and 8.
We compared our Rip-NeRF with the representative cutting-edge methods, \ie, NeRF~\cite{mildenhall2020nerf}, Mip-NeRF~\cite{barron2021mip},  TensoRF~\cite{chen2022tensorf}, Instant-NGP~\cite{muller2022instant}, Tri-MipRF~\cite{hu2023Tri-MipRF}, Zip-NeRF~\cite{barron2023zipnerf}, and 3D Gaussian Splatting (denoted as 3DGS)~\cite{kerbl3Dgaussians}.
Since some methods are not optimized for captures with variable distances or multiple resolutions, we incorporate the area loss $\mathcal{L}_\text{area}$ with all the methods for a fair comparison.
All the methods are retrained on the combined training and validation splits and evaluated on the testing split, using their official implementations, in alignment with Tri-MipRF~\cite{hu2023Tri-MipRF}.

\paragraph{Quantitative results.}
The quantitative results are presented in Tab.~\ref{tab:avg_multiblender_results}, where we assess rendering quality using PSNR, SSIM~\cite{wang2004image}, and VGG LPIPS~\cite{zhang2018unreasonable} metrics.
Additionally, to evaluate the efficiency of computation and storage, we report the average training time on an NVIDIA A100-SXM4-80GB GPU and the model size.
We can see that methods without anti-aliasing design, \ie, NeRF, TensoRF, Instant-NGP, and 3DGS, perform poorly in the multi-scale setting, which is the consequence of the discrete sampling in the continuous physical space.
Importantly, our Rip-NeRF consistently outperforms all the other cutting-edge methods, even the strong Tri-MipRF and Zip-NeRF baselines, in terms of PSNR and LPIPS metrics.
Though Zip-NeRF performs slightly better in terms of the SSIM metric, it requires almost double training time (4.5h \vs 2.6h) and four times of model size (592 MB \vs 160 MB).
Notably, the variant of our method that reduces the training iterations from $120k$ to $25k$, Rip-NeRF$_\text{25k}$, also achieves comparable results to Zip-NeRF, while requiring only 11.76\% training time (4.5h \vs 32min).
Additionally, the GPU memory consumption during training of Zip-NeRF is about 80 GB, which is unfeasible for some consumer-level GPUs, \eg the NVIDIA GeForce RTX 4090 \etc, while that of Rip-NeRF is only 20 GB.
And, the rendering speed of our Rip-NeRF is about 3 FPS in this experimental setting, which outperforms Zip-NeRF (0.25 FPS).
Admittedly, Zip-NeRF is designed for the more general unbounded scenes and we believe its performance can be further optimized for bounded objects. Nevertheless, our work demonstrates the effectiveness of area-sampling for high-fidelity anti-aliased rendering.
The effectiveness, efficiency, and compactness of our Rip-NeRF are attributed to our Ripmap-Encoded Platonic solid representation, which enables efficient anisotropic area-sampling of the 3D space.
In contrast, the multi-sampling mechanism in Zip-NeRF is effective but not efficient enough, which inherently demands a large number of samples to featurize a single area, putting it in a dilemma between the rendering quality and efficiency of computation and storage.

\paragraph{Qualitative results.}
To qualitatively evaluate the performance of our Rip-NeRF, we compare it with Tri-MipRF~\cite{hu2023Tri-MipRF} and Zip-NeRF~\cite{barron2023zipnerf}, since they are the most relevant methods to our approach and perform well quantitatively in the multi-scale setting.
We compared the full-resolution renderings of the three methods in Fig.~\ref{fig:qualitative_multiscale} and the first row of Fig.~\ref{fig:teaser} (teaser), where we can see that our Rip-NeRF achieves the best rendering quality, particularly in regions with challenging appearance and geometry, such as the anisotropic specular highlights on the gong in the ``drums'' scene, the slender rope on the ``ship'' scene, and the periodic grids on the ``mic'' scene.
Besides evaluating full-resolution renderings, we also present the renderings of the three methods at $\nicefrac{1}{2}$, $\nicefrac{1}{4}$, and $\nicefrac{1}{8}$ resolutions in Fig.~\ref{fig:qualitative_aliasing}, to demonstrate the effectiveness of our Rip-NeRF in anti-aliasing and preserving fine details.
We can see that in scenarios with lower resolution, our Rip-NeRF method exhibits superior performance compared to the other two methods. Notably, the periodic features on the microphone, which appear blurred in images rendered by Tri-MipRF and exhibit aliasing (``jaggies'') in those by Zip-NeRF, are rendered with higher fidelity in our results. These qualitative findings further corroborate the efficacy of Rip-NeRF in producing high-fidelity, anti-aliasing images, demonstrating its robustness across various resolution settings.

%-----------------------------------------------------------------------------
\begin{table}[]
    \renewcommand{\tabcolsep}{6pt}
    \centering
    \resizebox{\linewidth}{!}{
        \begin{tabular}{@{}l|ccc@{\,}}
             & \!PSNR $\uparrow$\! & \!SSIM $\uparrow$\! & \!LPIPS $\downarrow$
            \\
            \hline
            SRN~\cite{srn} & 22.26  & 0.846  & 0.170 \\
LLFF~\cite{mildenhall2019local} & 24.88 & 0.911 & 0.114 \\
Neural Volumes~\cite{neuralvolumes} & 26.05 & 0.893 & 0.160 \\
% Plenoxels~\cite{fridovich2022plenoxels} & 31.71 & 0.958 & 0.049 \\
NeRF~\cite{mildenhall2020nerf} & 31.74 &  0.953 & 0.050 \\
DVGO~\cite{sun2022direct} & 31.95 & 0.957 & 0.053 \\
TensoRF~\cite{chen2022tensorf} & 33.14 & 0.963 & 0.047 \\
Instant-NGP~\cite{muller2022instant} & 33.18 & 0.963 & 0.045 \\
3DGS~\cite{kerbl3Dgaussians} & 34.44 & \cellcolor{orange}0.975 & \cellcolor{red}0.028 \\

\hline
Mip-NeRF~\cite{barron2021mip} & 33.09 & 0.961 & 0.043 \\
Tri-MipRF ~\cite{hu2023Tri-MipRF} & 33.90 & 0.964 & 0.050 \\
Zip-NeRF~\cite{barron2023zipnerf} & \cellcolor{yellow}34.76 & \cellcolor{red}0.977 & \cellcolor{orange}0.032  \\
\hline
Rip-NeRF$_\text{25k}$ & \cellcolor{orange}34.92 & 0.969 & 0.042 \\
Rip-NeRF (Ours) & \cellcolor{red}35.44 & \cellcolor{yellow}0.973 & \cellcolor{yellow}0.037
        \end{tabular}
    }
    \vspace{0.5em}
    \caption{
        Quantitative performance on the single-scale Blender dataset~\cite{mildenhall2020nerf}. We compared our Rip-NeRF, and its variant, Rip-NeRF$_\text{25k}$, against several representative methods. The best, second-best, and third-best results are
        marked in red, orange, and yellow, respectively.
    }
    \vspace{-2.5em}
    \label{tab:avg_blender_results}
\end{table}
%-----------------------------------------------------------------------------

\begin{table*}[]
    \renewcommand{\tabcolsep}{1pt}
    \centering
    \resizebox{\linewidth}{!}{
        \begin{tabular}{l|p{1cm}<{\centering}p{1cm}<{\centering}p{1cm}<{\centering}p{1cm}<{\centering}p{1cm}<{\centering}|p{1cm}<{\centering}p{1cm}<{\centering}p{1cm}<{\centering}p{1cm}<{\centering}p{1cm}<{\centering}|p{1cm}<{\centering}p{1cm}<{\centering}p{1cm}<{\centering}p{1cm}<{\centering}p{1cm}<{\centering}p{1cm}<{\centering}}
            & \multicolumn{5}{c}{PSNR$\uparrow$} & \multicolumn{5}{c}{SSIM$\uparrow$} & \multicolumn{5}{c}{LPIPS$\downarrow$} \\

& \textit{lego} & \textit{flask} & \textit{mic} & \textit{filter} & \textit{Avg} & \textit{lego} & \textit{flask} & \textit{mic} & \textit{filter} & \textit{Avg} & \textit{lego} & \textit{flask} & \textit{mic} & \textit{filter} & \textit{Avg} \\ \hline

Zip-NeRF~\cite{barron2023zipnerf}& 35.95&37.79 &39.15 &38.45 &37.84 &0.991 & 0.992&0.997 &\cellcolor{red}0.991 &\cellcolor{red}0.993 &0.014 & 0.012&0.008 &0.021 &0.014 \\

Tri-MipRF~\cite{hu2023Tri-MipRF}& 37.19 &37.38 & 40.30&36.58 &38.09 & 0.991& 0.991&0.996 &0.985 &0.991 & 0.012&0.013 &0.006&0.022&0.013 \\

Rip-NeRF (Ours)&\cellcolor{red}37.85&\cellcolor{red}38.46&\cellcolor{red}42.07&\cellcolor{red}37.17&\cellcolor{red}38.89 &\cellcolor{red}0.993&\cellcolor{red}0.993&\cellcolor{red}0.998&0.986&0.992&\cellcolor{red}0.010&\cellcolor{red}0.010&\cellcolor{red}0.004&\cellcolor{red}0.020&\cellcolor{red}0.011
        \end{tabular}
    }
    \vspace{0.3em}
    \caption{Quantitative results on our newly captured real-world dataset. The highest performance is marked in red.}
    \vspace{-2em}
    \label{tab:real}
\end{table*}

\begin{table*}[]
    \renewcommand{\tabcolsep}{1pt}
    \centering
    \resizebox{\linewidth}{!}{
        \begin{tabular}{l|c|p{1.1cm}<{\centering}p{1cm}<{\centering}p{1cm}<{\centering}p{1cm}<{\centering}p{1cm}<{\centering}|p{1.1cm}<{\centering}p{1cm}<{\centering}p{1cm}<{\centering}p{1cm}<{\centering}p{1cm}<{\centering}|p{1.1cm}<{\centering}p{1cm}<{\centering}p{1cm}<{\centering}p{1cm}<{\centering}p{1cm}<{\centering}}
            
 & & \multicolumn{5}{c}{PSNR$\uparrow$} & \multicolumn{5}{c}{SSIM$\uparrow$} & \multicolumn{5}{c}{LPIPS$\downarrow$} \\
 & Train $\downarrow$   & Full Res. & $\nicefrac{1}{2}$ Res. & $\nicefrac{1}{4}$ Res. & $\nicefrac{1}{8}$ Res. & Avg.  & Full Res. & $\nicefrac{1}{2}$ Res. & $\nicefrac{1}{4}$ Res. & $\nicefrac{1}{8}$ Res. & Avg.  & Full Res. & $\nicefrac{1}{2}$ Res. & $\nicefrac{1}{4}$ Res. & $\nicefrac{1}{8}$ Res. & Avg.  \\ \hline
 Tri-MipRF~\cite{hu2023Tri-MipRF} & \cellcolor{red}5.5 mins &
33.57 & 35.21 & 35.96 & 36.46 & 35.30 &
0.962 & \cellcolor{yellow}0.975 & \cellcolor{yellow}0.982 & \cellcolor{yellow}0.987 & \cellcolor{yellow}0.976 &
0.052 & \cellcolor{yellow}0.029 & \cellcolor{yellow}0.019 & \cellcolor{yellow}0.013 & \cellcolor{yellow}0.028 \\
Rip-NeRF, PS3 (w/o PSP)       & \cellcolor{yellow}25.0 mins  & 33.32     & 34.65    & 35.25    & 35.92    & 34.79 & 0.959     & 0.972    & 0.979    & 0.985    & 0.974 & 0.056     & 0.032    & 0.021    & 0.014    & 0.031 \\  
Rip-NeRF, PS4                 & 25.5 mins  & 32.76     & 34.14    & 34.63    & 34.63    & 34.06 & 0.950     & 0.961    & 0.964    & 0.970    & 0.961 & 0.066     & 0.042    & 0.031    & 0.028    & 0.041 \\
Rip-NeRF, PS6                 & 26.5 mins & \cellcolor{yellow}33.86     & \cellcolor{yellow}35.37    & \cellcolor{yellow}36.34    & \cellcolor{yellow}36.78    & \cellcolor{yellow}35.59 & \cellcolor{yellow}0.963     & 0.974    & \cellcolor{yellow}0.982    & 0.986    & \cellcolor{yellow} 0.976 & \cellcolor{yellow}0.051     &\cellcolor{yellow} 0.029    & \cellcolor{yellow}0.019    & \cellcolor{yellow}0.013    & \cellcolor{yellow}0.028 \\ 
Rip-NeRF w/o RE               & \cellcolor{orange} 8.5 mins & 33.64     & 35.26    & 36.13    & 36.68    & 35.43 & 0.962     & 0.974    & 0.981    & \cellcolor{yellow}0.987    & \cellcolor{yellow}0.976 & 0.052     & 0.030    & 0.020    & \cellcolor{yellow}0.013    & 0.029 \\ 
Rip-NeRF$_\text{25k}$ & 32.0 mins & \cellcolor{orange}34.30     & \cellcolor{orange}35.94    & \cellcolor{orange}36.92    & \cellcolor{orange}37.47    & \cellcolor{orange}36.16 & \cellcolor{orange}0.966     & \cellcolor{orange}0.978    & \cellcolor{orange}0.984    & \cellcolor{orange}0.989    & \cellcolor{orange}0.979 & \cellcolor{orange}0.045     & \cellcolor{orange}0.025    & \cellcolor{orange}0.016    & \cellcolor{orange}0.011    & \cellcolor{orange}0.024 \\ 
Rip-NeRF (Ours)               & 2.6 hrs & \cellcolor{red}35.30     & 
 \cellcolor{red}37.01    & \cellcolor{red}38.07    & \cellcolor{red}38.54    & \cellcolor{red}37.23 & \cellcolor{red}0.973     & \cellcolor{red}0.983    & \cellcolor{red}0.988    & \cellcolor{red}0.991    & \cellcolor{red}0.984 & \cellcolor{red}0.037     & \cellcolor{red}0.019    & \cellcolor{red}0.011    & \cellcolor{red}0.008    & \cellcolor{red}0.019  

        \end{tabular}
    }
    \vspace{0.3em}
    \caption{Quantitative Comparison of Rip-NeRF and its ablations on the Multi-Scale Blender Dataset~\cite{barron2021mip}. The best, second-best, and third-best results are marked in red, orange, and yellow, respectively.}
    \vspace{-2em}
    \label{tab:multi_scale_ablation}
\end{table*}

\subsection{Evaluation on the Single-scale Blender Dataset}
The single-scale Blender dataset~\cite{mildenhall2020nerf} renders the image at a fixed resolution and distance, which is consistent with the assumption of methods without anti-aliasing design, \eg, NeRF~\cite{mildenhall2020nerf}, TensoRF~\cite{chen2022tensorf}, Instant-NGP~\cite{muller2022instant}, and 3DGS~\cite{kerbl3Dgaussians}.
Even though this scenario is not our primary focus, we still compared our Rip-NeRF and Rip-NeRF$_\text{25k}$ on this dataset against several representative methods.
The quantitative results of this comparison are detailed in Tab.~\ref{tab:avg_blender_results}, where we can observe that our Rip-NeRF and Rip-NeRF$_\text{25k}$ perform the best and second-best, respectively, in terms of the PSNR metric, while achieving comparable results to the Zip-NeRF and 3DGS in terms of the SSIM and LPIPS metrics.
It demonstrates that our Rip-NeRF is also effective in the single-scale setting, even though it is not optimized for this scenario.

\subsection{Evaluation on Real-world Captures}
To further verify the practicality of our approach, we applied our Rip-NeRF to a newly captured real-world dataset.
We captured four challenging objects that contain fine periodic structures using an iPhone.
We applied Structure-from-Motion (SfM) to the image sequence to estimate camera parameters and then employed image segmentation to separate the targets from the background scene.
We also applied a rigid transform and rescaling to the poses reconstructed by SfM, aiming to center and scale the target object in a manner similar to that used in the Blender dataset.
Each captured scene consists of 400 to 420 images with a resolution of 3840 $\times$ 2160, down-sampled with factors of 2, 4, 8 and 16 (instead of 1, 2, 4, and 8 in Multi-scale Blender) due to its relatively high original resolution.
For reconstruction, we uniformly sampled 50\% of the images, reserving the remaining portion for evaluation.
This setup makes the number of images and resolution of two splits similar to the previous Multi-scale Blender Dataset. However, noisy camera poses, motion blur, and defocus blur imply greater challenges to reconstruction than synthetic data.

We compared our Rip-NeRF with Tri-MipRF~\cite{hu2023Tri-MipRF} and Zip-NeRF~\cite{barron2023zipnerf} since they show strong capabilities in the previous experiments.
Three example results are shown in Fig.~\ref{fig:qualitative_real}, where we can clearly observe that our Rip-NeRF renders more accurate intricate structures and appearance details.
Additionally, quantitative results of the whole dataset in Tab.~\ref{tab:real} and the PSNR/SSIM values displayed below each image further affirm the effectiveness of our approach for real-world captures. 

\subsection{Ablation study}
\label{ablation_study}

In our ablation studies, we evaluated key components of our method. Rip-NeRF w/o PSP utilized orthogonal triplanes with anisotropic area sampling to assess the impact of excluding Platonic Solid Projection. For examining the role of anisotropic area sampling, Rip-NeRF w/o RE employed isotropic mipmaps, adapting the Tri-MipRF~\cite{hu2023Tri-MipRF} framework with 10 unparalleled planes from an icosahedron, using nvdiffrast~\cite{Laine2020diffrast} for mipmap construction. The two ablation experiments are both trained for 25000 iterations.

The quantitative results from our experiments on the multi-scale Blender dataset are detailed in Table \ref{tab:multi_scale_ablation}. Our findings reveal that the independent application of each component—Platonic Solid Projection (PSP) and Ripmap Encoding (RE) —yields only modest improvements. Specifically, employing PSP alone results in a mere 0.37\% increase in average PSNR, while using solely RE slightly lowers all three metrics. However, when PSP and RE are combined in our Rip-NeRF$_{25k}$ model, they produce a synergistic effect: an average 3.44\% increase in PSNR, with SSIM and LPIPS significantly surpassing those of Tri-MipRF~\cite{hu2023Tri-MipRF}. This synergy suggests that the whole is greater than the sum of its parts. While PSP alone improves alignment of the ellipsoid footprint with the 2D grid, it still relies on square query areas. Ripmap Encoding with only 3 planes are not capable of modeling the nuance of different Gaussians as well.

In our ablation study focusing on the choice of Platonic Solid, we experimented with cubes (PS3), tetrahedrons (PS4), and dodecahedron (PS6). In this ablation experiment, we conducted experiment by increasing grid resolution for PS3, PS4, and PS6 to keep the total parameter count the same as our full method (160MB). From PS3 to our full model (PS10), the training time increases 28\%. Interestingly, adding planes does not always correlate with performance improvements. For instance, despite the addition of one plane in PS4 compared to PS3, there was a decrease in PSNR, SSIM, and LPIPS. Conversely, applying more planes to the tri-mip encoding method yielded minimal enhancement, suggesting that ripmaps overcome model capacity limitations. This is evidenced by the fact that isotropic spheres project similarly onto different planes, whereas anisotropic 3D Gaussian footprints vary across planes.

\subsection{Limitations}
Despite the excellent performance on bounded datasets, our representation still faces challenges for unbounded scenes. We suppose two possible reasons that cause difficulties in 2D-style representations. First, non-vaguely-convex shapes lead to information from self-occluded locations being projected onto the same 2D area. Second, the space warping mechanism proposed in Mip-NeRF360 \cite{barron2022mip} encourages more locations along a non-linear curve, compared to a straight line, to be projected onto the same 2D area, which is difficult for the explicit 2D feature grid to characterize. To address these challenges, perhaps a more advanced 3D to 2D mapping function is required to be explored.
\section{Conclusion}

In this work, we present a Ripmap-Encoded Platonic Solid representation for neural radiance fields, named Rip-NeRF.
Our Rip-NeRF can render high-fidelity anti-aliasing images while maintaining efficiency, enabled by the proposed Platonic Solid Projection and Ripmap Encoding.
The Platonic Solid Projection factorizes the 3D space onto the unparalleled faces of a certain Platonic solid, such that the anisotropic 3D areas can be projected onto planes with distinguishable characterization.
And the Ripmap Encoding enables featurizing the projected anisotropic areas both precisely and efficiently by the anisotropic area-sampling.
These two components work together for precisely and efficiently featurizing anisotropic 3D areas.
It achieves state-of-the-art rendering quality on both synthetic datasets and real-world captures, particularly excelling in the fine details of structures and textures, which verifies the effectiveness of the proposed Platonic Solid Projection and Ripmap Encoding.

% \clearpage

\bibliographystyle{ACM-Reference-Format}
\bibliography{main.bib}

\appendix

%-----------------------------------------------------------------------------
\begin{figure*}[!t]
    \centering
    \includegraphics[width=0.95\textwidth]{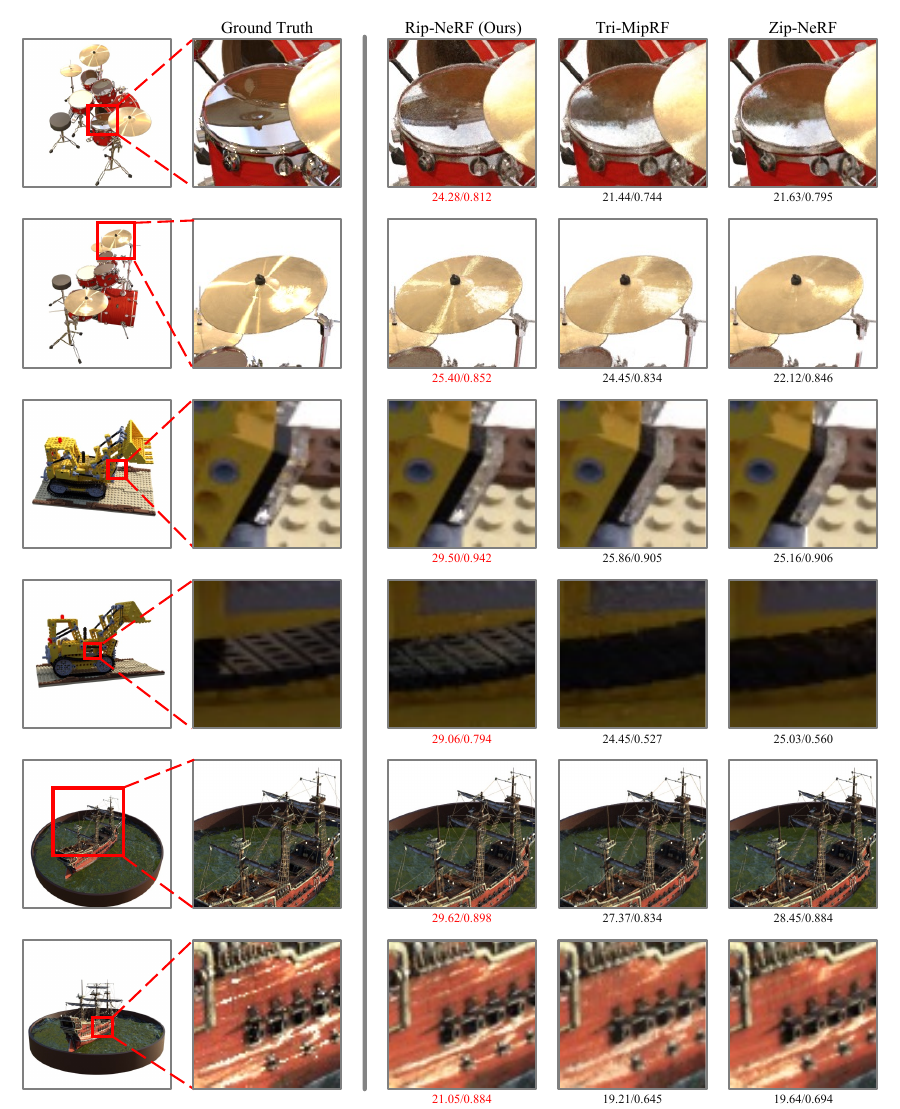}
    \Description{Qualitative comparison of the full-resolution (close-up views) renderings on the multi-scale Blender dataset. More details are presented by our method.}
    \vspace{-1.0em}
    \caption{
        Qualitative comparison of the full-resolution (close-up views) renderings on the multi-scale Blender dataset. PSNR/SSIM values are shown at the bottom of each result.
    }
    \vspace{-1em}
    \label{fig:qualitative_multiscale}
\end{figure*}
%-----------------------------------------------------------------------------

%-----------------------------------------------------------------------------
\begin{figure*}[!t]
    \centering
    \includegraphics[width=\linewidth]{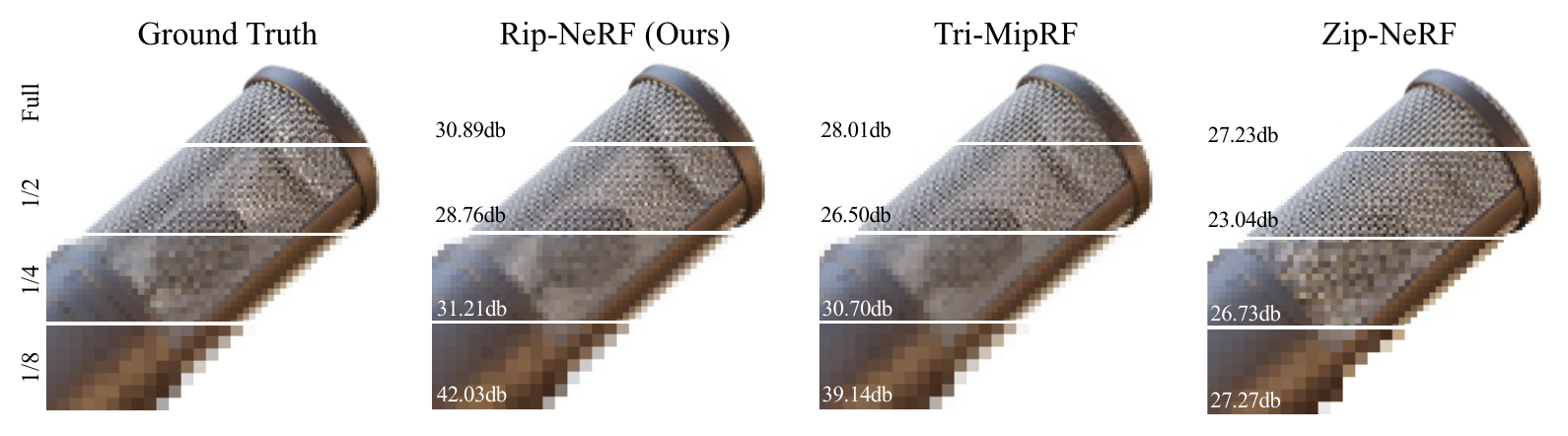}
    \Description{Qualitative comparison of the multi-resolution evaluation on the mic scene from the multi-scale Blender dataset. Our method can capture more high-frequency details on different scales.}
    \vspace{-8mm}
    \caption{
        Qualitative comparison of the multi-resolution evaluation on the mic scene from the multi-scale Blender dataset. PSNR values are shown in the bottom right corners of each result.
    }
    % \vspace{0em}
    \label{fig:qualitative_aliasing}
\end{figure*}
%-----------------------------------------------------------------------------

%-----------------------------------------------------------------------------
\begin{figure*}[!t]
    \centering
    \includegraphics[width=\linewidth]{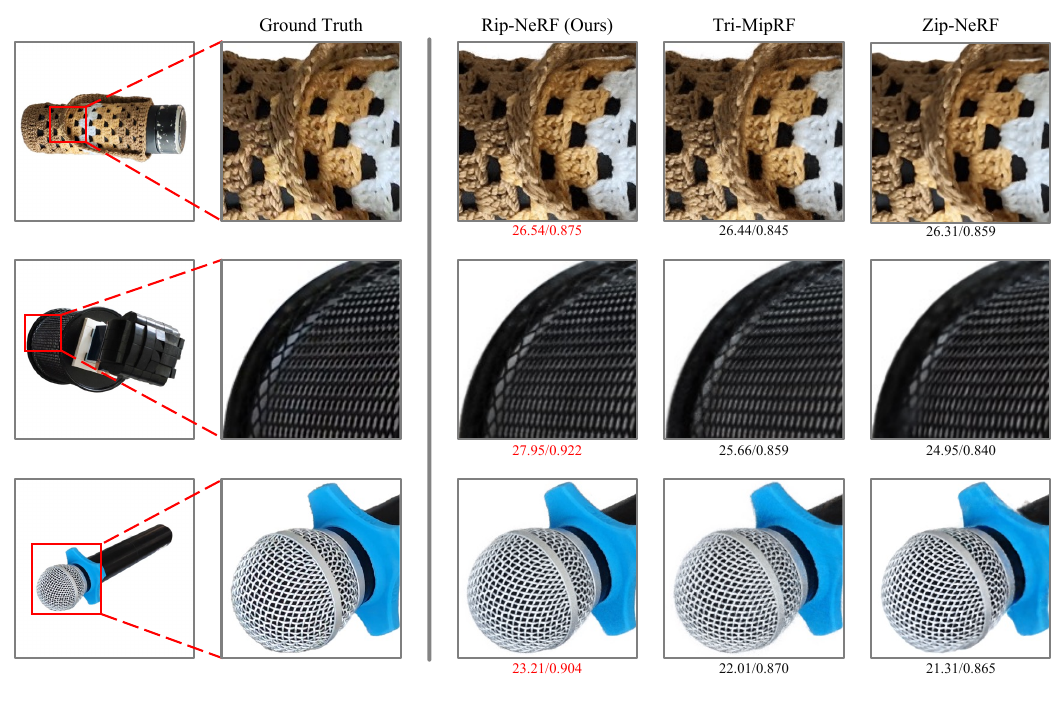}
    \Description{Qualitative comparison on our real-world dataset. More details can be captured on the images our model produces.}
    \vspace{-8mm}
    \caption{
        Qualitative results of our Rip-NeRF, Tri-MipRF~\cite{hu2023Tri-MipRF}, and Zip-NeRF~\cite{barron2023zipnerf} on real-world captures. PSNR/SSIM values are shown at the bottom of each result.
    }
    % \vspace{-4em}
    \label{fig:qualitative_real}
\end{figure*}
%-----------------------------------------------------------------------------

\clearpage

\appendix
\section{Cone Casting}
To characterize an anisotropic sub-volume, we follow the procedure introduced in Mip-NeRF~\cite{barron2021mip} to model the conical frustum as a 3D Gaussian $\mathcal{N}(\boldsymbol{\mu}, \boldsymbol{\Sigma})$, where $\boldsymbol{\mu}$ and $\boldsymbol{\Sigma}$ are the mean and covariance, respectively. Given a conical frustum defined by its near and far $\zval$ values $[\zval_0, \zval_1]$ along the ray direction $\raydir$ originating from the camera center $\rayorigin$, and the cone radius $\baseradius$ at the image plane, we can compute $\boldsymbol{\mu}$ and $\boldsymbol{\Sigma}$ as follows:

First, we calculate the mean $\mu_\zval$ and variance $\sigma_\zval^2$ of the frustum along the ray direction, as well as the variance $\sigma_{r}^2$ perpendicular to the ray direction:
\begin{align}
\mu_\zval = \frac{3\left(\zval_1^4-\zval_0^4\right)}{4\left(\zval_1^3-\zval_0^3\right)}, \;
\sigma_\zval^2 = \frac{3 \left(\zval_1^5 - \zval_0^5\right)}{5 \left( \zval_1^3 - \zval_0^3\right)} - \mu_\zval^2, \;
\sigma_{r}^2 = \baseradius^2 \left(\frac{3 \left( \zval_1^5 - \zval_0^5 \right)}{20 \left(\zval_1^3 - \zval_0^3\right)}\right).
\end{align}

Then, we transform these quantities from the local coordinate frame of the conical frustum to the world coordinate frame, obtaining the mean $\boldsymbol{\mu}$ and covariance $\boldsymbol{\Sigma}$ of the 3D Gaussian:

\begin{equation}
\boldsymbol{\mu} = \rayorigin + \mu_\zval \raydir, \quad
\boldsymbol{\Sigma} = \sigma_\zval^2 \left(\raydir\raydir^\transpose\right) + \sigma_{r}^2 \left(\mathbf{I} - \frac{\raydir\raydir^\transpose}{\norm{\raydir}_2^2} \right),
\end{equation}
where $\mathbf{I}$ is the identity matrix. The resulting 3D Gaussian $\mathcal{N}(\boldsymbol{\mu}, \boldsymbol{\Sigma})$ approximates the geometry of the conical frustum and serves as the input for the subsequent Platonic Solid Projection and Ripmap Encoding steps in Rip-NeRF.

\section{Tetra-linear Interpolation}
The tetra-linear interpolation process in Ripmap Encoding involves querying the feature values from the surrounding vertices in the 4D space defined by the position $(p_x, \; p_y)$ and level $(l_x, \; l_y)$. Given a query point $(p_x, \; p_y, \; l_x, \; l_y)$, we first identify the 16 neighboring vertices in the 4D space. Let $(p_x^i, \; p_y^j, \; l_x^k, \; l_y^m)$ denote the neighboring vertex, where $i, j, k, m \in \{0, 1\}$ represent the binary indices in each dimension. The feature value at the query point is then interpolated using the weighted sum of the feature values at these neighboring vertices:

\begin{equation}
\begin{aligned}
\label{eq:tetra-linear-interpolation}
\boldsymbol{f} = \sum_{i=0}^1 \sum_{j=0}^1 \sum_{k=0}^1 \sum_{m=0}^1 w_{ijkm} \cdot \boldsymbol{f}_{ijkm},
\end{aligned}
\end{equation}
where $\boldsymbol{f}_{ijkm} = \mipmap^{l_x^k, \; l_y^m}(p_x^i, \; p_y^j)$ represents the feature value at the neighboring vertex $(p_x^i, \; p_y^j, \; l_x^k, \; l_y^m)$, and $w_{ijkm}$ is the interpolation weight calculated based on the distances between the query point and the neighboring vertex in each dimension:

\begin{equation}
\begin{aligned}
w_{ijkm} = \prod_{\substack{d \in \{x, y\} \\ t \in \{p, l\}}} (1 - |t_d - t_d^{ij}|),
\end{aligned}
\end{equation}
where $t_d^{ij}$ represents the neighboring vertex coordinates along dimension $d$, with $i$ and $j$ being the binary indices for position and level, respectively. By performing this tetra-linear interpolation, we obtain a smooth and continuous feature representation $\boldsymbol{f}$ for the anisotropic 2D Gaussian at the query point $(p_x, \; p_y, \; l_x, \; l_y)$ in the Ripmap Encoding. This interpolation process allows for efficient querying of the Ripmap Encoding while considering the anisotropic nature of the Gaussian footprint.

\section{Implementation Details}

\subsection{Tiny MLP}
\label{subsec:mlp}
Our tiny MLP nonlinearly maps the Ripmap-Encoding feature vector $\feat$ and the view direction $\viewdir$ to the density $\density$ and color $\col$ of the sampled sphere $\sphere$, aligning with Tri-MipRF~\cite{hu2023Tri-MipRF} for consistency. The dimension of $\feat$ is 160, considering the ripmaps $\mipmap$ shape of $512 \times 512 \times 16$ and the use of 10 unparalleled icosahedron faces. The first two MLP layers process $\feat$ to yield $\density$ and a 15-dimensional geometric feature $\feat_\text{geo}$. The view direction $\viewdir$, encoded via spherical harmonics, is combined with $\feat_\text{geo}$ in the final three layers to estimate the view-dependent color $\col$, similar to~\cite{muller2022instant}. The MLP's width is set to 128, using ReLU activations (except for the output layer of $\density$, where a truncated exponential function is used, following~\cite{muller2022instant}). This efficient MLP, implemented with tiny-cuda-nn, is optimized for fused and half-precision operations.

\subsection{Sampling Strategy}
Sample points are selected within a sphere of radius $r$, predefined for each scene. For the nerf-synthetic dataset scenes $r$ is set to 1.5. We adopt the OccupancyGrid from nerfacc~\cite{li2022nerfacc} as our sampler, setting density to zero for points outside the bounding sphere. The occupancy threshold was 0.005 for all experiments.

\subsection{Optimization}
Optimizable parameters in Rip-NeRF include the tiny MLP's model weights and the ripmaps $\mipmap$. Model weights are initialized following~\cite{glorot2010understanding}, while ripmaps $\mipmap$ start from a uniform distribution over $[-0.01, 0.01]$. We use AdamW~\cite{loshchilov2019decoupled} for optimization, scaling the base learning rate for $\mipmap$ by 10x due to its direct scene representation role. The base learning rate is $2 \times 10^{-3}$, reduced by 0.6x at steps 60K, 90K, 100K, and 108K, over a total of 120K iterations. Following~\cite{muller2022instant}, we dynamically adjust the batch size to maintain approximately 256K spheres per batch.

%-----------------------------------------------------------------------------

%-----------------------------------------------------------------------------
\section{Detailed Quantitative Results}

For a more detailed quantitative per-scene analysis, Table \ref{tab:avg_multiscale_per_scene_results} and Table \ref{tab:avg_singlescale_per_scene_results} showcases our Rip-NeRF against representative baseline methods training on the multi-scale and single-scale blender dataset.

\begin{table*}[!p]
    \renewcommand{\tabcolsep}{1pt}
    \centering
    \resizebox{0.9\linewidth}{!}{
    \begin{tabular}{l|p{1.3cm}<{\centering}p{1.3cm}<{\centering}p{1.3cm}<{\centering}p{1.3cm}<{\centering}p{1.3cm}<{\centering}p{1.3cm}<{\centering}p{1.3cm}<{\centering}p{1.3cm}<{\centering}|p{1.3cm}<{\centering}}
    & \multicolumn{9}{c}{\textbf{PSNR} $\uparrow$}                                                     \\
& \textit{chair} & \textit{drums} & \textit{ficus} & \textit{hotdog} & \textit{lego}  & \textit{materials} & \textit{mic}   & \textit{ship}  & \textit{Average} \\ \hline
NeRF w/o $\mathcal{L}_\text{area}$ & 29.92 & 23.27 & 27.15 & 32.00  & 27.75 & 26.30     & 28.40 & 26.46 & 27.66   \\
NeRF~\cite{mildenhall2020nerf}                               & 33.39 & 25.87 & 30.37 & 35.64  & 31.65 & 30.18     & 32.60 & 30.09 & 31.23   \\
Mip-NeRF~\cite{barron2021mip}                            & 37.14 & 27.02 & 33.19 & 39.31  & 35.74 & 32.56     & 38.04 & 33.08 & 34.51   \\ \hline
TensoRF~\cite{chen2022tensorf}                            & 32.47 & 25.37 & 31.16 & 34.96  & 31.73 & 28.53     & 31.48 & 29.08 & 30.60   \\
Instant-NGP~\cite{muller2022instant}                        & 32.95 & 26.43 & 30.41 & 35.87  & 31.83 & 29.31 & 32.58 & 30.23 & 31.20   \\
% Zip-NeRF~\cite{barron2023zipnerf}                           & 25.34 & 20.21 & 24.31 & 28.56  & 24.79 & 23.33     & 24.75 & 23.74 & 24.38   \\
Tri-MipRF~\cite{hu2023Tri-MipRF}                          & 38.36 & 28.66 & 34.29 & 40.02 & 36.57 & 32.22     & 38.46 & 33.80 & 35.30   \\ 
Zip-NeRF~\cite{barron2023zipnerf}& \cellcolor{orange}39.52 & \cellcolor{orange}29.46 & \cellcolor{yellow}35.54 & \cellcolor{orange}41.64 & \cellcolor{orange}37.25 & \cellcolor{red}34.39 & \cellcolor{yellow}39.69 & \cellcolor{orange}36.08 & \cellcolor{orange}36.70 \\
3DGS~\cite{kerbl3Dgaussians}                               & 33.04 & 25.52 & 29.21 & 35.48  & 29.66 & 27.25 & 31.15 & 29.12 & 30.05   \\ \hline
Rip-NeRF, PS3 (w/o PSP) & 37.64& 28.39 & 33.58 & 39.45  & 35.98 & 31.91     & 38.11 & 33.20 & 34.79   \\
Rip-NeRF, PS4 & 37.73 & 28.56 & 34.58 & 38.85  & 34.50 & 31.66     & 38.29 & 28.28 & 34.06   \\
Rip-NeRF, PS6 & 38.80 & 29.07 & 35.27 & 40.25 & 36.27 & 32.20     & 39.32 & 33.55 & 35.59   \\
Rip-NeRF, w/o PE & 38.99 & 28.92 & 35.21 & 39.97  & 35.87 & 32.32 & 38.92 & 33.24 & 35.43   \\
Rip-NeRF$_\text{25k}$ & \cellcolor{yellow}39.36 & \cellcolor{yellow}29.40  & \cellcolor{orange}35.89 & \cellcolor{yellow}40.82  & \cellcolor{yellow}36.94 & \cellcolor{yellow}32.66 & \cellcolor{orange}40.07 & \cellcolor{yellow}34.13 & \cellcolor{yellow}36.16   \\
Rip-NeRF (ours) & \cellcolor{red}40.08 & \cellcolor{red}29.88  & \cellcolor{red}36.61 & \cellcolor{red}41.98  & \cellcolor{red}38.36 & \cellcolor{orange}33.79 & \cellcolor{red}40.94 & \cellcolor{red}36.18 & \cellcolor{red}37.23   \\

% %%%%%%%%%%%%%%%%%%%%%%%%%%%%%%%%%%%%%%%%%%%%%%%%%%%%%%%%%%%%
\multicolumn{9}{c}{} \\
& \multicolumn{9}{c}{\textbf{SSIM} $\uparrow$}                                                     \\
& \textit{chair} & \textit{drums} & \textit{ficus} & \textit{hotdog} & \textit{lego}  & \textit{materials} & \textit{mic}   & \textit{ship}  & \textit{Average} \\ \hline
NeRF w/o $\mathcal{L}_\text{area}$ & 0.944 & 0.891 & 0.942 & 0.959  & 0.926 & 0.934     & 0.958 & 0.861 & 0.927   \\
NeRF~\cite{mildenhall2020nerf}                               & 0.971 & 0.932 & 0.971 & 0.979  & 0.965 & 0.967     & 0.980 & 0.900 & 0.958   \\
MipNeRF~\cite{barron2021mip}                            & 0.988 & 0.945 & 0.984 & 0.988  & 0.984 & \cellcolor{yellow}0.977     & 0.993 & 0.922 & 0.973   \\ \hline
TensoRF~\cite{chen2022tensorf}                            & 0.967 & 0.930 & 0.974 & 0.977  & 0.967 & 0.957     & 0.978 & 0.895 & 0.956   \\
Instant-ngp~\cite{muller2022instant}                        & 0.971 & 0.940 & 0.973 & 0.979  & 0.966 & 0.959     & 0.981 & 0.904 & 0.959   \\
Tri-MipRF~\cite{hu2023Tri-MipRF} & 0.991 & 0.957 & 0.986 & \cellcolor{yellow}0.990  & \cellcolor{yellow}0.987 & 0.972     & 0.993 & 0.936 & 0.976   \\ 
Zip-NeRF~\cite{barron2023zipnerf}& \cellcolor{red}0.994 & \cellcolor{red}0.968 & \cellcolor{orange}0.991 & \cellcolor{red}0.993 & \cellcolor{orange}0.990 & \cellcolor{red}0.984 & \cellcolor{orange}0.995 & \cellcolor{red}0.966 & \cellcolor{red}0.985 \\
3DGS~\cite{kerbl3Dgaussians} & 0.979 & 0.944 & 0.970 & 0.984  & 0.967 & 0.959     & 0.979 & 0.922 & 0.963   \\ \hline
Rip-NeRF, PS3 (w/o PSP) & 0.989 & 0.955 & 0.984 & 0.988  & 0.984 & 0.969 & 0.992 & 0.928 & 0.974   \\
Rip-NeRF, PS4 & 0.984 & 0.951 & 0.987 & 0.983  & 0.976 & 0.953     & 0.988 & 0.868 & 0.961   \\
Rip-NeRF, PS6 & \cellcolor{yellow}0.992 & 0.959 & 0.989 & \cellcolor{yellow}0.990  & 0.984 & 0.972 & \cellcolor{yellow}0.994 & 0.930 & 0.976   \\
Rip-NeRF, w/o PE & \cellcolor{yellow}0.992 & 0.959 & 0.989 & 0.989  & 0.983 & 0.973     & 0.993 & 0.929 & 0.976   \\
Rip-NeRF$_\text{25k}$ & \cellcolor{orange}0.993 & \cellcolor{yellow}0.963 & \cellcolor{yellow}0.990 & \cellcolor{orange}0.991  & \cellcolor{yellow}0.987 & 0.976     & \cellcolor{orange}0.995 & \cellcolor{yellow}0.939 & \cellcolor{yellow}0.979   \\
Rip-NeRF (ours) & \cellcolor{red}0.994 &\cellcolor{orange}0.966	& \cellcolor{red}0.992	& \cellcolor{red}0.993	&\cellcolor{red}0.991	&\cellcolor{orange}0.980	&\cellcolor{red}0.996	&\cellcolor{orange}0.960	&\cellcolor{orange}0.984   \\

% %%%%%%%%%%%%%%%%%%%%%%%%%%%%%%%%%%%%%%%%%%%%%%%%%%%%%%%%%%%%
\multicolumn{9}{c}{} \\
& \multicolumn{9}{c}{\textbf{LPIPS} $\downarrow$}                                                    \\
& \textit{chair} & \textit{drums} & \textit{ficus} & \textit{hotdog} & \textit{lego}  & \textit{materials} & \textit{mic}   & \textit{ship}  & \textit{Average} \\ \hline
NeRF w/o $\mathcal{L}_\text{area}$ & 0.035 & 0.069 & 0.032 & 0.028  & 0.041 & 0.045     & 0.031 & 0.095 & 0.052   \\
NeRF~\cite{mildenhall2020nerf} & 0.028 & 0.059 & 0.026 & 0.024  & 0.035 & 0.033 & 0.025 & 0.085 & 0.044   \\
MipNeRF~\cite{barron2021mip}                            & \cellcolor{yellow}0.011 & 0.044 & 0.014 & \cellcolor{orange}0.012  & \cellcolor{orange}0.013 & \cellcolor{red}0.019     & \cellcolor{orange}0.007 & \cellcolor{yellow}0.062 & 0.026   \\ \hline
TensoRF~\cite{chen2022tensorf} & 0.042 & 0.075 & 0.032 & 0.035  & 0.036 & 0.063     & 0.040 & 0.112 & 0.054   \\
Instant-ngp~\cite{muller2022instant}                        & 0.035 & 0.066 & 0.029 & 0.028  & 0.040 & 0.051     & 0.032 & 0.095 & 0.047   \\
Tri-MipRF~\cite{hu2023Tri-MipRF}                          & 0.013 & 0.049 & 0.017 & 0.016  & \cellcolor{yellow}0.014 & 0.036     & 0.010 & 0.071 & 0.028   \\
Zip-NeRF~\cite{barron2023zipnerf} & 0.017 & \cellcolor{orange}0.038 & \cellcolor{red}0.010 & \cellcolor{yellow}0.013 & \cellcolor{orange}0.013 & \cellcolor{red}0.019 & \cellcolor{yellow}0.008 & \cellcolor{orange}0.049 & \cellcolor{orange}0.021 \\
3DGS~\cite{kerbl3Dgaussians}                               & 0.024 & 0.056 & 0.029 & 0.020  & 0.037 & 0.039     & 0.024 & 0.082 & 0.039   \\  \hline
Rip-NeRF, PS3 (w/o PSP)              & 0.015 & 0.051 & 0.020 & 0.019  & 0.016 & 0.039     & 0.012 & 0.077 & 0.031   \\
Rip-NeRF, PS4                      & 0.015 & 0.052 & 0.016 & 0.023  & 0.028 & 0.042     & 0.010 & 0.146 & 0.042   \\
Rip-NeRF, PS6                      & 0.012 & 0.046 & \cellcolor{yellow}0.013 & 0.015  & 0.017 & 0.036     & 0.009 & 0.077 & 0.028   \\
Rip-NeRF, w/o PE                  & \cellcolor{yellow}0.011 & 0.046 & \cellcolor{yellow}0.013 & 0.017  & 0.018 & 0.034 & 0.009 & 0.081 & 0.029   \\
Rip-NeRF$_\text{25k}$ & \cellcolor{orange}0.010 & \cellcolor{yellow}0.041 & \cellcolor{orange}0.011 & \cellcolor{yellow}0.013  & \cellcolor{orange}0.013 & \cellcolor{yellow}0.029     & \cellcolor{orange}0.007 & 0.069 & \cellcolor{yellow}0.024 \\
Rip-NeRF (ours) &\cellcolor{red}0.009	&\cellcolor{red}0.037	&\cellcolor{red}0.010	&\cellcolor{red}0.010	&\cellcolor{red}0.009	&\cellcolor{orange}0.024	&\cellcolor{red}0.006	&\cellcolor{red}0.047	&\cellcolor{red}0.019
    \end{tabular}
    }
    \caption{Quantitative Per-Scene Results on the Multi-Scale Blender Dataset: This table presents the arithmetic mean of each metric, averaged over the four scales of the dataset for individual scenes. Performance rankings are highlighted with color coding: the best, second-best, and third-best results are in red, orange, and yellow, respectively.}
    \label{tab:avg_multiscale_per_scene_results}
\end{table*}

\begin{table*}[!p]
    \renewcommand{\tabcolsep}{1pt}
    \centering
    \resizebox{0.9\linewidth}{!}{
    \begin{tabular}{l|p{1.3cm}<{\centering}p{1.3cm}<{\centering}p{1.3cm}<{\centering}p{1.3cm}<{\centering}p{1.3cm}<{\centering}p{1.3cm}<{\centering}p{1.3cm}<{\centering}p{1.3cm}<{\centering}|p{1.3cm}<{\centering}}
                          & \multicolumn{9}{c}{\textbf{PSNR} $\uparrow$}                                                   \\
                      & \textit{chair} & \textit{drums} & \textit{ficus} & \textit{hotdog} & \textit{lego}  & \textit{materials} & \textit{mic}   & \textit{ship}  & \textit{Average}   \\
\hline
SRN~\cite{srn} & 29.96 & 17.18 & 20.73 & 26.81 & 20.85 & 18.09 & 26.85 & 20.60 & 22.26 \\
LLFF~\cite{mildenhall2019local} & 28.72 & 21.13 & 21.79 & 31.41 & 24.54 & 20.72 & 27.48 & 23.22 & 24.88 \\
Neural Volumes~\cite{neuralvolumes} & 28.33 & 22.58 & 24.79 & 30.71 & 26.08 & 24.22 & 27.78 & 23.93 & 26.05 \\
NeRF~\cite{mildenhall2020nerf} & 34.17 & 25.08 & 30.39 & 36.82 & 33.31 & 30.03 & 34.78 & 29.30 & 31.74 \\
Mip-NeRF~\cite{barron2021mip}               & 35.14 & 25.48 & 33.29 & 37.48  & 35.70 & 30.71     & 36.51 & 30.41 & 33.09 \\
\hline
Mip-NeRF 360~\cite{barron2022mip} & 35.65 & 25.60 & 33.19 & 37.71  & 36.10 & 29.90 & 36.52 & 31.26 & 33.24 \\
DVGO~\cite{sun2022direct} & 34.09 & 25.44 & 32.78 & 36.74 & 34.64 & 29.57 & 33.20 & 29.13 & 31.95 \\
TensoRF~\cite{chen2022tensorf}               & 35.76 & 26.01 & 33.99 & 37.41  & 36.46 & 30.12     & 34.61 & 30.77 & 33.14 \\
Instant-NGP~\cite{muller2022instant}           & 35.00 & 26.02 & 33.51 & 37.40  & 36.39 & 29.78     & 36.22 & 31.10 & 33.18 \\
Tri-MipRF~\cite{hu2023Tri-MipRF} & \cellcolor{yellow}36.53 & 26.73 & 34.93 & 38.46  & \cellcolor{yellow}36.61 & 30.63     & \cellcolor{yellow}37.56 & 29.74 & 33.90 \\
Zip-NeRF~\cite{barron2023zipnerf}  & 36.19 & \cellcolor{yellow}27.37 & 36.08 & \cellcolor{orange}39.18 & 36.60 & \cellcolor{red}32.63 & 36.74 & \cellcolor{red}33.27 & \cellcolor{yellow}34.76 \\
3DGS~\cite{kerbl3Dgaussians} & 36.88 & 26.80 & \cellcolor{yellow}36.09 & 38.72  & 36.82 & 30.99 & 36.69 & \cellcolor{orange}32.52 & 34.44 \\
\hline
Rip-NeRF$_\text{25k}$ & \cellcolor{orange}37.31 & \cellcolor{orange}27.46 & \cellcolor{orange}36.45 & \cellcolor{yellow}39.04 & \cellcolor{orange}37.72 & \cellcolor{yellow}31.24 & \cellcolor{orange}39.28 & 30.88 & \cellcolor{orange}34.92 \\
Rip-NeRF (ours) & \cellcolor{red}37.58 & \cellcolor{red}27.76 & \cellcolor{red}36.87 & \cellcolor{red}39.77 & \cellcolor{red}38.29 & \cellcolor{orange}31.80 & \cellcolor{red}39.91 & \cellcolor{yellow}31.56 & \cellcolor{red}35.44 \\

\multicolumn{9}{c}{} \\
                      & \multicolumn{9}{c}{\textbf{SSIM} $\uparrow$}                                                   \\
                      & \textit{chair} & \textit{drums} & \textit{ficus} & \textit{hotdog} & \textit{lego}  & \textit{materials} & \textit{mic}   & \textit{ship}  & \textit{Average}   \\
\hline
SRN~\cite{srn} & 0.910 & 0.766 & 0.849 & 0.923 & 0.809 & 0.808 & 0.947 & 0.757 & 0.846 \\ 
LLFF~\cite{mildenhall2019local} & 0.948 & 0.890 & 0.896 & 0.965 & 0.911 & 0.890 & 0.964 & 0.823 & 0.911 \\
Neural Volumes~\cite{neuralvolumes} & 0.916 & 0.873 & 0.910 & 0.944 & 0.880 & 0.888 & 0.946 & 0.784 & 0.893 \\
NeRF~\cite{mildenhall2020nerf} & 0.975 & 0.925 & 0.967 & 0.979 & 0.968 & 0.953 & 0.987 & 0.869 & 0.953 \\
Mip-NeRF~\cite{barron2021mip} & 0.981 & 0.932 & 0.980 & 0.982  & 0.978 & 0.959 & 0.991 & 0.882 & 0.961 \\
\hline
Mip-NeRF 360~\cite{barron2022mip} & 0.983 & 0.931 & 0.979 & 0.982  & 0.980 & 0.949     & 0.991 & 0.893 & 0.961 \\
DVGO~\cite{sun2022direct} & 0.977 & 0.930 & 0.978 & 0.980 & 0.976 & 0.951 & 0.983 & 0.879 & 0.957 \\
TensoRF~\cite{chen2022tensorf} & 0.985 & 0.937 & 0.982 & 0.982  & 0.983 & 0.952 & 0.988 & 0.895 & 0.963 \\
Instant-NGP~\cite{muller2022instant} & 0.979 & 0.937 & 0.981 & 0.982  & 0.982 & 0.951 & 0.990 & 0.896 & 0.963 \\
Tri-MipRF~\cite{hu2023Tri-MipRF} & 0.987 & 0.940 & 0.984 & \cellcolor{yellow}0.984  & 0.983 & 0.952 & \cellcolor{yellow}0.992 & 0.886 & 0.964 \\
Zip-NeRF~\cite{barron2023zipnerf} & \cellcolor{yellow}0.988 & \cellcolor{orange}0.957 & \cellcolor{orange}0.990 & \cellcolor{red}0.988 & \cellcolor{yellow}0.985 & \cellcolor{red}0.974 & \cellcolor{orange}0.993 & \cellcolor{red}0.945 & \cellcolor{red}0.977 \\ 
3DGS~\cite{kerbl3Dgaussians} & \cellcolor{red}0.990 & \cellcolor{red}0.961 & \cellcolor{red}0.991 & \cellcolor{red}0.988 & \cellcolor{orange}0.986 & 0.969 & 0.993 & \cellcolor{orange}0.918 & \cellcolor{orange}0.975 \\
\hline
Rip-NeRF$_\text{25k}$ & \cellcolor{orange}0.989 & 0.948 & \cellcolor{yellow}0.989 & \cellcolor{orange}0.986 & \cellcolor{yellow}0.985 & \cellcolor{yellow}0.960 & \cellcolor{red}0.995 & \cellcolor{yellow}0.898 & 0.969 \\
Rip-NeRF (ours) & \cellcolor{red}0.990 & \cellcolor{yellow}0.950 & \cellcolor{orange}0.990 & \cellcolor{red}0.988 & \cellcolor{red}0.987 & \cellcolor{orange}0.964 & \cellcolor{red}0.995 & \cellcolor{orange}0.918 & \cellcolor{yellow}0.973 \\

\multicolumn{9}{c}{} \\
                      & \multicolumn{9}{c}{\textbf{LPIPS} $\downarrow$}                                                  \\
                      & \textit{chair} & \textit{drums} & \textit{ficus} & \textit{hotdog} & \textit{lego}  & \textit{materials} & \textit{mic}   & \textit{ship}  & \textit{Average}   \\
\hline
SRN~\cite{srn} & 0.106 & 0.267 & 0.149 & 0.100 & 0.200 & 0.174 & 0.063 & 0.299 & 0.170 \\
LLFF~\cite{mildenhall2019local} & 0.064 & 0.126 & 0.130 & 0.061 & 0.110 & 0.117 & 0.084 & 0.218 & 0.114 \\
Neural Volumes~\cite{neuralvolumes} & 0.109 & 0.214 & 0.162 & 0.109 & 0.175 & 0.130 & 0.107 & 0.276 & 0.160 \\
NeRF~\cite{mildenhall2020nerf} & 0.026 & 0.071 & 0.032 & 0.030 & 0.031 & 0.047 & 0.012 & 0.150 & 0.050 \\
Mip-NeRF~\cite{barron2021mip} & 0.021 & 0.065 & 0.020 & 0.027  & 0.021 & \cellcolor{yellow}0.040     & 0.009 & 0.138 & 0.043 \\
\hline
Mip-NeRF 360~\cite{barron2022mip} & 0.018 & 0.069 & 0.022 & 0.024  & 0.018 & 0.053     & 0.011 & 0.135 & 0.042 \\
DVGO~\cite{sun2022direct} & 0.027 & 0.077 & 0.024 & 0.034 & 0.028 & 0.058 & 0.017 & 0.161 & 0.053 \\ 
TensoRF~\cite{chen2022tensorf}               & 0.022 & 0.073 & 0.022 & 0.032  & 0.018 & 0.058     & 0.015 & 0.138 & 0.047 \\
Instant-NGP~\cite{muller2022instant}           & 0.022 & 0.071 & 0.023 & 0.027  & 0.017 & 0.060 & 0.010 & 0.132 & 0.045 \\
Tri-MipRF~\cite{hu2023Tri-MipRF} & 0.021 & 0.076 & 0.025 & 0.031  & 0.019 & 0.067     & 0.012 & 0.148 & 0.050 \\
Zip-NeRF~\cite{barron2023zipnerf} & \cellcolor{orange}0.016 & \cellcolor{orange}0.048 & \cellcolor{orange}0.012 & \cellcolor{orange}0.019 & 0.017 & \cellcolor{orange}0.036 & \cellcolor{yellow}0.007 & \cellcolor{red}0.099 & \cellcolor{orange}0.032 \\ 
3DGS~\cite{kerbl3Dgaussians} & \cellcolor{red}0.010 & \cellcolor{red}0.035 & \cellcolor{red}0.009 & \cellcolor{red}0.018 & \cellcolor{red}0.013 & \cellcolor{red}0.029 & \cellcolor{red}0.005 & \cellcolor{orange}0.104 & \cellcolor{red}0.028 \\
\hline
Rip-NeRF$_\text{25k}$ & \cellcolor{yellow}0.017 & 0.064 & 0.016 & 0.027 & \cellcolor{yellow}0.016 & 0.053 & 0.008 & 0.135 & 0.042 \\
Rip-NeRF (ours) & \cellcolor{orange}0.016 & \cellcolor{yellow}0.062 & \cellcolor{yellow}0.015 & \cellcolor{yellow}0.022 & \cellcolor{orange}0.014 & 0.046 & \cellcolor{orange}0.006 & \cellcolor{yellow}0.111 & \cellcolor{yellow}0.037 \\
    \end{tabular}
    }
    \caption{Quantitative Per-Scene Results on the Single-Scale Blender Dataset: This table presents the arithmetic mean of each metric, averaged over the four scales of the dataset for individual scenes. Performance rankings are highlighted with color coding: the best, second-best, and third-best results are in red, orange, and yellow, respectively.
    }
    \label{tab:avg_singlescale_per_scene_results}
\end{table*}

\section{More qualitative Results}
\begin{figure*}[!b]
    \centering
    \includegraphics[width=0.73\linewidth]{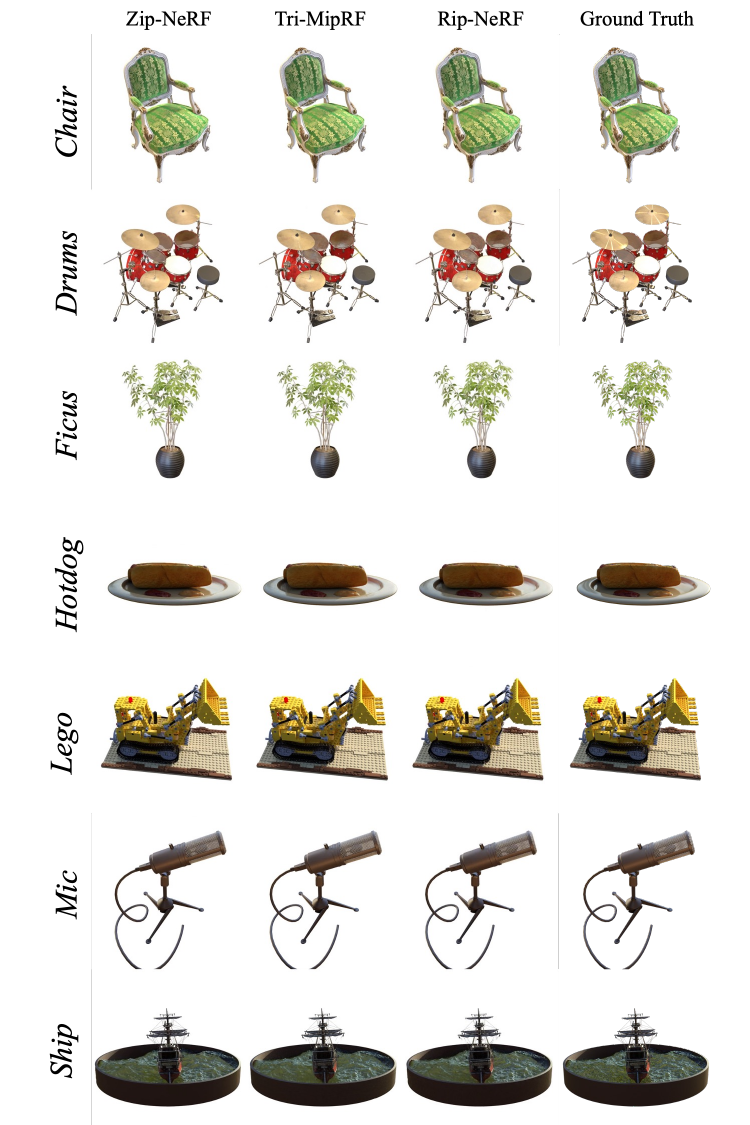}
    \Description{More qualitative rendering results of Zip-NeRF, Tri-MipRF, and our Rip-NeRF on the multi-scale Blender dataset.}
    % \vspace{-1mm}
    \caption{
        More qualitative rendering results of Zip-NeRF, Tri-MipRF, and our Rip-NeRF on the multi-scale Blender dataset.
    }
    % \vspace{0em}
    \label{fig:supp_rgb}
\end{figure*}
%-----------------------------------------------------------------------------

\end{document}